%% file: iclr2027_conference.tex
\documentclass{article}

\usepackage{iclr2027_conference,times}
\input{math_commands.tex}

\usepackage{amsmath,amssymb,amsfonts}
\usepackage{booktabs}
\usepackage{multirow}
\usepackage{tabularx}
\usepackage{longtable}
\usepackage{graphicx}
\usepackage{adjustbox}
\usepackage{capt-of}
\usepackage{float}
\usepackage{wrapfig}
\usepackage[table]{xcolor}
\usepackage[most]{tcolorbox}
\usepackage{enumitem}
\usepackage{fontawesome5}
\usepackage{microtype}
\usepackage{url}
\usepackage[colorlinks=true,allcolors=blue]{hyperref}
\usepackage{xspace}

\newcommand{\dataname}{SpaceCast-Bench\xspace}

\tcbset{
  aibox/.style={
    width=\linewidth,
    top=8pt,
    bottom=4pt,
    left=6pt,
    right=6pt,
    colback=blue!6!white,
    colframe=black,
    colbacktitle=black,
    coltitle=white,
    fonttitle=\bfseries,
    fontupper=\small,
    enhanced,
    breakable,
    attach boxed title to top left={yshift=-0.1in,xshift=0.15in},
    boxed title style={boxrule=0pt,colframe=white},
  }
}
\newtcolorbox{AIbox}[2][]{aibox,title={#2},#1}
\newcommand{\promptslot}[1]{\textcolor{red}{\textit{\{#1\}}}}
\newcommand{\promptangle}[1]{\textless #1\textgreater}

\definecolor{takeawayblue}{HTML}{4F86E8}
\definecolor{takeawaybg}{HTML}{EEF5FD}
\newtcolorbox{keytakeaways}[1]{
    enhanced,
    colback=takeawaybg,
    colframe=takeawayblue,
    boxrule=0pt,
    leftrule=2.2pt,
    arc=2mm,
    left=6pt,
    right=6pt,
    top=10pt,
    bottom=4pt,
    before skip=9pt,
    after skip=9pt,
    attach boxed title to top left={xshift=8pt,yshift=-2mm},
    boxed title style={
        colback=takeawayblue,
        colframe=takeawayblue,
        boxrule=0pt,
        arc=1.4mm,
        left=4pt,
        right=4pt,
        top=1pt,
        bottom=1pt
    },
    coltitle=white,
    fonttitle=\small\bfseries,
    fontupper=\small\itshape,
    title={\faLightbulb\ Key Takeaways: #1}
}


\title{\dataname : Evaluating Predictive Spatial Reasoning in Vision-Language Models}

\author{%
  \textbf{Hongxing Li}$^{1}$%
  \thanks{Equal contribution.\quad $^{\dagger}$Corresponding author.}, ~
  \textbf{Jinyue Su}$^{1}$\footnotemark[1], ~
  \textbf{Dingming Li}$^{1}$, ~
  \textbf{Wenqi Zhang}$^{1}$, ~
  \textbf{Weiming Lu}$^{1}$, ~ 
  \textbf{Jun Xiao}$^{1}$, ~ \\
  \textbf{Yueting Zhuang}$^{1}$, ~
  \textbf{Yongliang Shen}$^{1}$\footnotemark[2] \\[6pt]
  $^{1}$Zhejiang University \\
  \texttt{\{hongxing.li, syl\}@zju.edu.cn} \\
  \\
  \href{https://zjureal.com/SpaceCast-Bench}{\raisebox{-0.15em}{\includegraphics[height=0.95em]{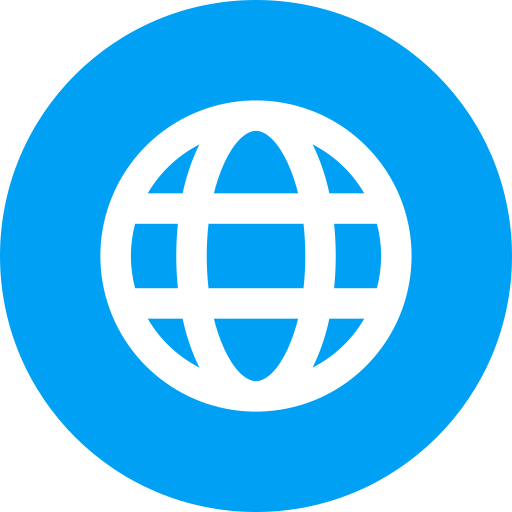}}\hspace{0.25em}Website}
  \hspace{1.5em}
  \href{https://github.com/ZJU-REAL/SpaceCast-Bench}{\raisebox{-0.15em}{\includegraphics[height=0.95em]{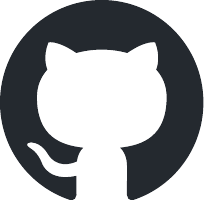}}\hspace{0.25em}Code}
  \hspace{1.5em}
  \href{https://huggingface.co/datasets/hongxingli/SpaceCast-Bench}{\raisebox{-0.15em}{\includegraphics[height=0.95em]{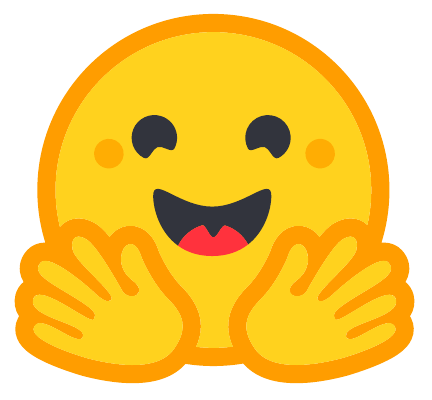}}\hspace{0.25em}Dataset}
}

\iclrfinalcopy

\begin{document}

\maketitle
\lhead{}
\renewcommand{\headrulewidth}{0pt}

\input{sections/abstract}

\input{sections/introduction}

\input{sections/related_work}

\input{sections/methodology}

\input{sections/experiments}

\input{sections/conclusion}
\input{sections/statement}

\bibliographystyle{iclr2027_conference}
\bibliography{iclr2027_conference}

\appendix
\input{sections/appendix}

\end{document}

%% file: math_commands.tex
\usepackage{amsmath,amsfonts,bm}

\def\eqref#1{equation~\ref{#1}}

\def\1{\bm{1}}

\DeclareMathAlphabet{\mathsfit}{\encodingdefault}{\sfdefault}{m}{sl}
\SetMathAlphabet{\mathsfit}{bold}{\encodingdefault}{\sfdefault}{bx}{n}

%% file: sections/abstract.tex
\begin{abstract}
Existing spatial reasoning benchmarks mainly test spatial perception: reading off relations already visible in the input. Yet real-world spatial intelligence demands \textbf{predictive spatial reasoning}: constructing a scene from observations, anticipating how an intervention changes it, and reasoning about the unseen outcome. We introduce \textbf{\dataname{}}, the first benchmark to directly and diagnostically evaluate this capability. Built around an observe--transform--infer framework, its 3,862 questions from 182 real-world scenes span 16 task types at three levels: static perception, local prediction, and global prediction, progressively requiring scene understanding, spatial state updating, and relational inference over unobserved outcomes. Evaluating 21 models exposes a stark gap: the strongest model reaches only 58.0\% against 87.2\% human performance, while spatially specialized models remain near random chance. Controlled analyses further reveal that bridge views are critical for integrating distributed observations, and that explicit 3D evidence benefits models more reliably than generated outcome images or videos. Fine-tuning on our programmatically generated data lifts Qwen3-VL-4B from 34.0\% to 65.7\% with macro-average gains across six out-of-domain benchmarks.
\end{abstract}

%% file: sections/introduction.tex
\begin{figure}[H]
    \centering
    \includegraphics[width=\textwidth]{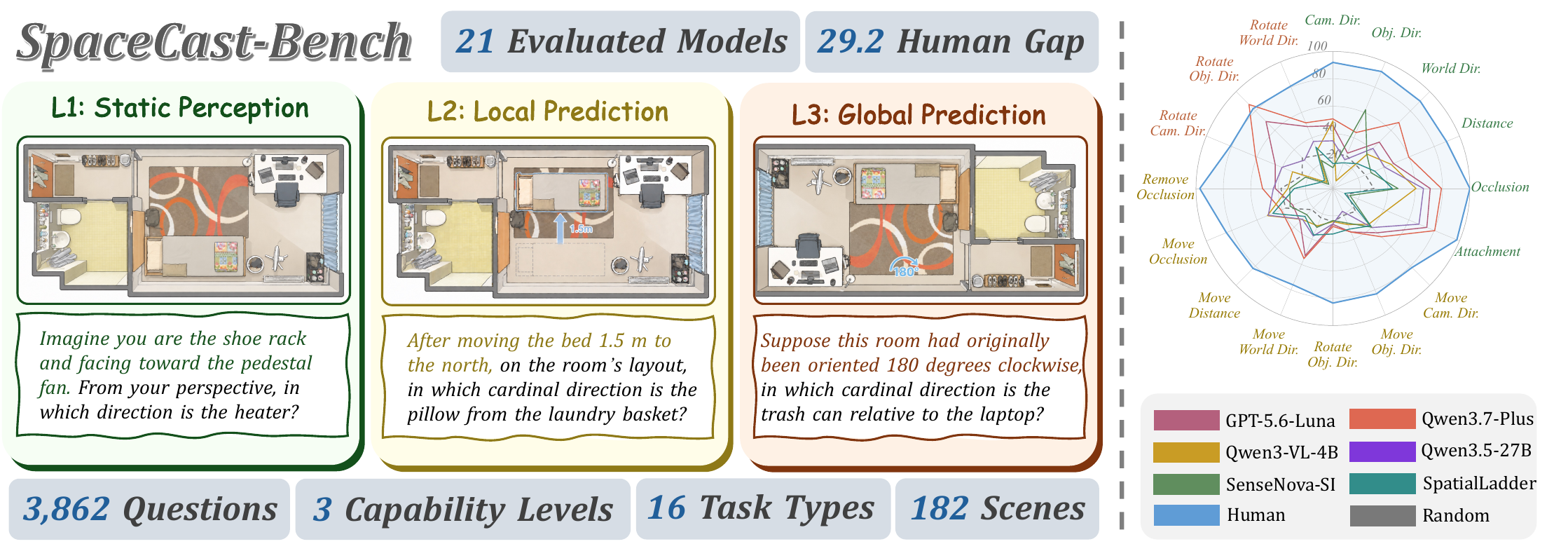}
    \vspace{-5pt}
    \caption{We introduce \dataname, a benchmark for evaluating predictive spatial reasoning. \textbf{Left:} representative tasks at three capability levels. \textbf{Right:} task-wise performance of six representative models, with human and random baselines.}
    \vspace{-5pt}
    \label{fig:teaser}
\end{figure}

\section{Introduction}

Visual-spatial intelligence entails perceiving and mentally manipulating spatial relationships, encompassing relational reasoning and the ability to transform between egocentric and allocentric perspectives~\citep{yang2025thinking,gardner2011frames}. This capacity is fundamental for vision-language models to understand and act in the physical world. Recent benchmarks spanning single-image relation grounding~\citep{kamath2023whats,du2024embspatial-bench}, 3D metric estimation~\citep{ma20253dsrbench,cheng2024spatialrgpt}, and multi-view integration~\citep{li2025viewspatial-bench,yang2026mmsi-bench,yang2025thinking} have driven rapid progress in spatial perception~\citep{ouyang2025spacer,wu2025spatial-mllm,li2026spatialladder}.

Yet these benchmarks share a fundamental limitation: they evaluate \textbf{spatial perception}, the ability to read off relations already present in the observed frames~\citep{tong2024cambrian,cheng2024spatialrgpt,zhang2025flatland,yang2025thinking}. What they overlook is \textbf{spatial prediction}, the ability to anticipate how a scene will be reconfigured after an intervention. Consider a robot retrieving an envelope beneath a book with a cup resting on top: to plan safely, it must predict that sliding the book will displace the cup and decide to move the cup first. This requires constructing a mental model of the scene, simulating the consequences of an action, and revising the plan accordingly. Such predictive spatial reasoning is a prerequisite for safe manipulation and multi-step planning, and precisely what spatial world models must learn~\citep{ha2018world,hafner2019learning,hafner2025mastering,assran2025v-jepa2}: represent an environment in a form that supports anticipating action consequences even when those consequences are never shown. Despite its centrality to embodied intelligence, no existing benchmark evaluates this capability in a controlled and diagnostically decomposable way.

To fill this gap, we introduce \textbf{\dataname}, the first benchmark to directly and diagnostically evaluate predictive spatial reasoning, comprising 3,862 questions from 182 scenes across 16 task types (Figure~\ref{fig:teaser}). Built around an observe--transform--infer framework, the benchmark progressively evaluates scene understanding, spatial state updating, and relational inference over unobserved outcomes. To instantiate this at scale, we develop a geometry-grounded pipeline that builds question-conditioned, depth-verified view chains from posed RGB-D scans, applies controlled local or global transformations in 3D, and derives programmatic labels while keeping the resulting state unobserved. Tasks span three capability levels: \textbf{L1 (Static Perception)}, probing spatial relations in an unmodified scene from single- or multi-view observations; \textbf{L2 (Local Prediction)}, testing anticipation of relation changes after localized object-level interventions; and \textbf{L3 (Global Prediction)}, testing reasoning after a rigid transformation is applied to the entire layout. Across the benchmark, tasks cover direction, distance, visibility and occlusion, and attachment structure, with directional relations posed in camera-centric, object-centric, and world-centric reference frames, enabling granular diagnosis of where model reasoning breaks down.

Experiments across 21 proprietary, open-weight, and spatially specialized models reveal a substantial gap: the strongest model achieves only 58.0\% accuracy against human performance of 87.2\%, while spatially specialized models remain near random chance, indicating that existing spatial training does not transfer to spatial state updating. Controlled analyses show that CoT~\citep{wei2022cot} prompting provides no consistent benefit, bridge views aid integration of distributed observations, and explicit 3D evidence benefits strong general-purpose models far more reliably than generated outcome images or videos. Finally, post-training experiments show that benchmark-aligned training substantially improves Qwen3-VL-4B: staged SFT raises overall accuracy from 34.0\% to 65.7\% on \dataname{} and yields consistent gains across six out-of-domain benchmarks, confirming that our pipeline supports both systematic diagnosis and transferable model improvement.

Our main contributions are as follows: \textbf{(i)} We introduce \dataname, the first benchmark for diagnostic evaluation of predictive spatial reasoning, with 3,862 questions from 182 scenes across 16 task types. \textbf{(ii)} We develop a geometry-grounded pipeline that builds depth-verified view chains, applies controlled 3D transformations, and programmatically derives labels for unobserved outcome states. \textbf{(iii)} We evaluate 21 models with controlled analyses revealing a substantial human--model gap and distinct bottlenecks across model families, and show that training on our generated data improves both in-domain and six out-of-domain benchmarks.

%% file: sections/related_work.tex
\section{Related Work}

\subsection{Benchmarks for Perceptive Spatial Reasoning}

Early spatial benchmarks evaluate relations in a single observed image. What'sUp~\citep{kamath2023whats} and EmbSpatial-Bench~\citep{du2024embspatial-bench} focus on basic directional and embodied spatial questions, while SpatialRGPT-Bench~\citep{cheng2024spatialrgpt} and 3DSRBench~\citep{ma20253dsrbench} broaden the relation set using 3D annotations and natural scenes. Multi-image and video benchmarks extend this to sequential or multi-view inputs: SPAR-Bench~\citep{zhang2025flatland} covers depth, distance, cross-view matching, and spatial relations, ViewSpatial-Bench~\citep{li2025viewspatial-bench} evaluates spatial localization from multiple perspectives, MMSI-Bench~\citep{yang2026mmsi-bench} provides a human-curated evaluation spanning positional relationships, attributes, and multi-step reasoning, and VSI-Bench~\citep{yang2025thinking} evaluates spatial configurations and metric estimation from egocentric video. All these benchmarks center on spatial perception of existing scenes, without requiring any reasoning about how relations change after an intervention.

\subsection{Benchmarks for Predictive Spatial Reasoning}

Recent benchmarks have begun to evaluate reasoning over scene change. CLEVRER~\citep{yi2020clevrer} evaluates physical events and counterfactuals in synthetic collision videos, and SAT~\citep{ray2025sat} combines observed motion with hypothetical ego actions, with both grounded in observed evidence. MindCube~\citep{wang2026mindcube} studies what-if object motion from limited views, SpaMEM~\citep{liao2026spamem} tests belief revision after object operations, and SpatialWorld~\citep{gao2026spatialworld} evaluates agents that gather evidence and act in a closed loop. While these works touch on aspects of predictive reasoning, \dataname~is the first comprehensive benchmark to directly evaluate predictive spatial reasoning under controlled scene transformations.

%% file: sections/methodology.tex
\begin{figure*}[h!]
    \centering
    \includegraphics[width=\textwidth]{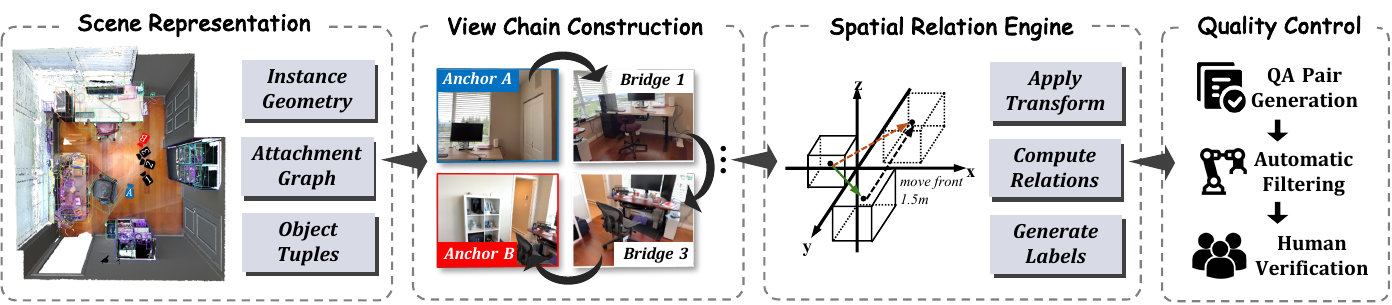}
    \vspace{-14pt}
    \caption{Overview of the \dataname construction pipeline.}
    \vspace{-10pt}
    \label{fig:construction-pipeline}
\end{figure*}

\section{\dataname}

\subsection{Overview}

\dataname evaluates whether a vision-language model can anticipate how spatial relations change after an intervention it never observes. The initial scene state $S_0$ is observed through an ordered sequence of RGB views $\mathcal{O} = (I_1, \ldots, I_m)$ sampled exclusively from the pre-transformation scene. Given $\mathcal{O}$, a natural-language transformation $T$, and a spatial query $q$, the model must evaluate $q$ in the unobserved result state $S_1 = F(S_0, T)$. This requires three latent operations: integrating $\mathcal{O}$ to recover the relevant portion of $S_0$, applying $T$ to obtain $S_1$, and evaluating $q$ in $S_1$. This observe--transform--infer pipeline distinguishes predictive spatial reasoning from the perceptual task of reading off relations already visible in the input.

\begin{figure*}[t]
    \centering
    \includegraphics[width=\textwidth]{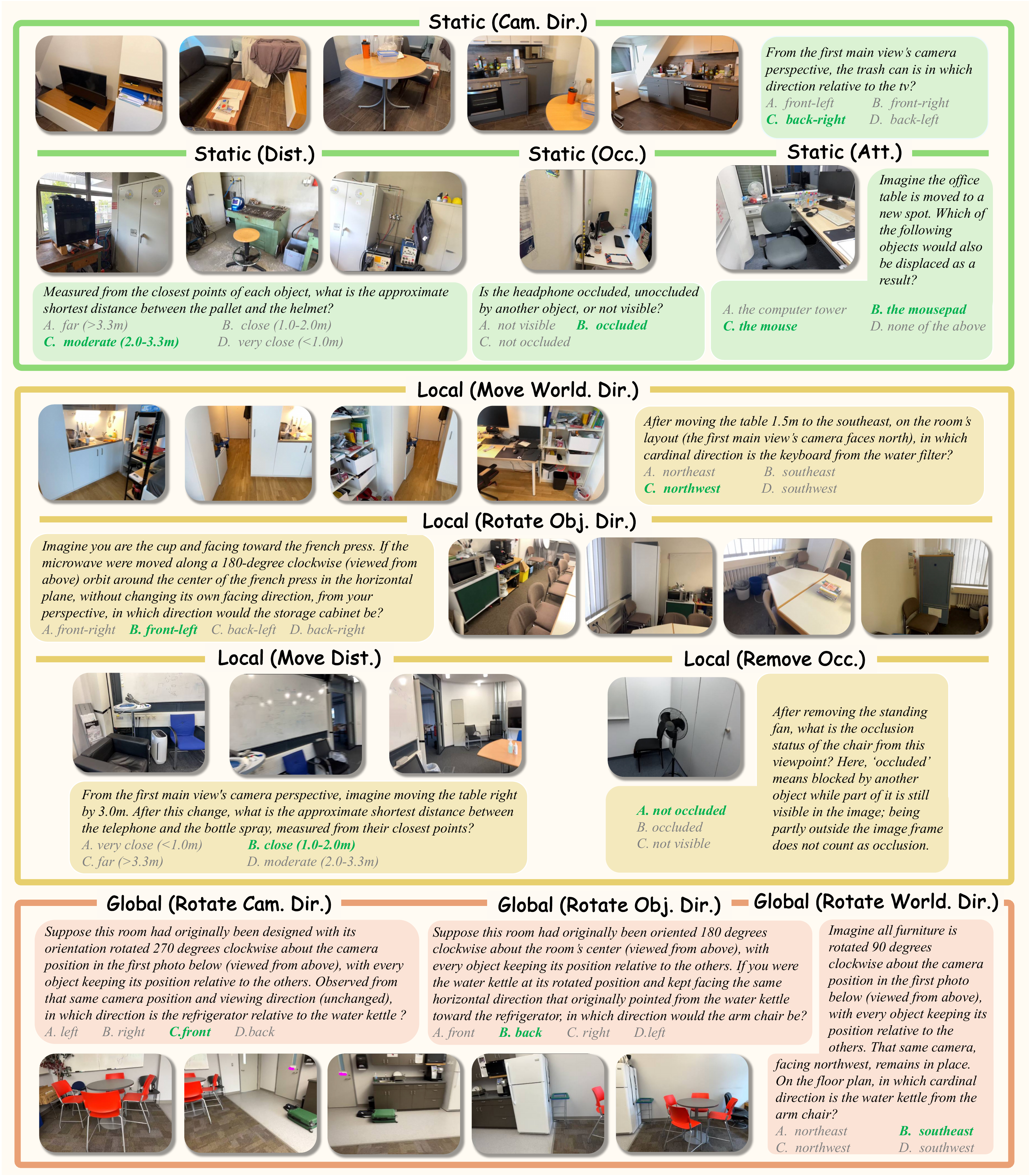}
    \vspace{-14pt}
    \caption{Representative tasks of \dataname. Correct answers are highlighted in green.}
    \vspace{-10pt}
    \label{fig:task-demonstration}
\end{figure*}

As illustrated by the representative task instances in Figure~\ref{fig:task-demonstration}, we organize tasks into three capability levels of increasing demand. \textbf{L1 (Static Perception)} sets $T$ to the identity, so $S_1 = S_0$ and the queried relation is directly recoverable from $\mathcal{O}$. These tasks probe spatial perception across direction, distance, visibility and occlusion, and attachment structure from single- or multi-view observations, establishing a perceptual baseline and isolating failures in the observe stage from those in the transform--infer stages. \textbf{L2 (Local Prediction)} applies a localized intervention to a single object, translating, orbiting, or removing it together with its attachment dependents, while leaving the rest of the scene intact. These tasks test whether the model can simulate an object-level action and evaluate the queried relation in the unobserved $S_1$ without corrupting its representation of unaffected regions. \textbf{L3 (Global Prediction)} applies a rigid yaw rotation to the complete layout, requiring a scene representation explicitly decoupled from the observed camera perspective and robust to conflation of scene rotation with a viewpoint change.

Across the benchmark, queried relations span direction, distance, visibility, occlusion, and attachment structure, with directional relations expressed in camera-centric, object-centric, and world-centric frames. This factorization across capability level, relation type, and reference frame enables targeted diagnosis of where model reasoning breaks down.

\subsection{Benchmark Construction}

\paragraph{Scene Representation.}
Figure~\ref{fig:construction-pipeline} summarizes our four-stage benchmark construction pipeline, built from 134 ScanNet~\citep{dai2017scannet} and 48 ScanNet++~\citep{yeshwanth2023scannet} scenes. Each scene provides RGB images with calibrated camera poses, depth maps, reconstructed 3D geometry, and instance-level semantic annotations; we derive each instance's surface geometry, 3D center, and axis-aligned bounding box from its mesh. We additionally construct a directed attachment graph where a parent-to-child edge indicates that displacing the parent also displaces the child; edges are proposed from surface contact, support overlap, and semantic compatibility, with each child assigned at most one parent. The transitive closure of this graph determines co-movement under local interventions and is used to generate attachment-chain questions. For each task type, we enumerate candidate object tuples and select a view chain grounded in their spatial arrangement.

\paragraph{View Chain Construction.}
Cross-view questions require two anchor views such that each anchor exposes its assigned query-object group while excluding the other. Candidate anchors are filtered by camera validity, image quality, projected object area, and geometric visibility, then selected to satisfy mutual semantic disjointness: neither anchor reveals the objects assigned to the other. Bridge views connect the two anchors via a depth-verified corridor between the aggregate centers of the two object groups; each consecutive pair must overlap the corridor by at least 10\% and advance toward the opposite anchor, and no bridge view may re-expose both groups. The resulting view chain is ordered along the selected traversal and contains only pre-transformation observations. Three task types require a single anchor; the remaining 13 use the disjoint-anchor construction.

\paragraph{Spatial Relation Engine.}
Given the object tuple and view chain, we compute relation labels directly from 3D geometry according to the task type. Direction projects object centers onto the floor plane and resolves their relative vector into eight horizontal sectors plus above and below; camera-centric, object-centric, and world-centric variants express this vector in the designated coordinate frame. Distance approximates the shortest surface-to-surface separation using sampled mesh points and maps it to four categorical bins of increasing distance. Visibility and occlusion use mesh ray casting from the calibrated camera center, with aggregated outcomes distinguishing not occluded, occluded, and not visible. Attachment queries the transitive closure of the attachment graph, requiring exact-set identification of all objects displaced when a root is moved. For predictive questions, we apply the specified transformation to the relevant object geometry, translating, orbiting, or removing an object or rotating the complete layout, and rerun the engine on the resulting state to obtain the ground-truth label. Geometrically ambiguous cases, labels near decision boundaries, and invalid motion endpoints are excluded across all relation families.

\paragraph{Quality Control.}
We combine automatic filtering with human verification to ensure geometric validity and question clarity. Automatic filtering removes geometrically ambiguous relations, labels near decision boundaries, invalid motion endpoints, insufficient visual grounding, and near-duplicates defined by scene, view, task type, and object identity. Answer positions are balanced within the generated pool before the benchmark is frozen. Human verification was conducted by 10 annotators, with every question receiving at least two independent reviews based on the model-visible input; reviewers checked object identifiability, visual answerability, and question clarity, with disagreements resolved by a third reviewer. Before adjudication, single-answer questions achieved 98.9\% pairwise agreement and attachment-chain questions achieved 91.8\% exact-set agreement. See Appendix~\ref{app:benchmark-details} for complete task definitions, construction details, and quality-control procedures.

\subsection{Dataset Statistics}
\begin{wrapfigure}[14]{r}{0.50\textwidth}
    \vspace{-15pt}
    \centering
    \includegraphics[width=\linewidth]{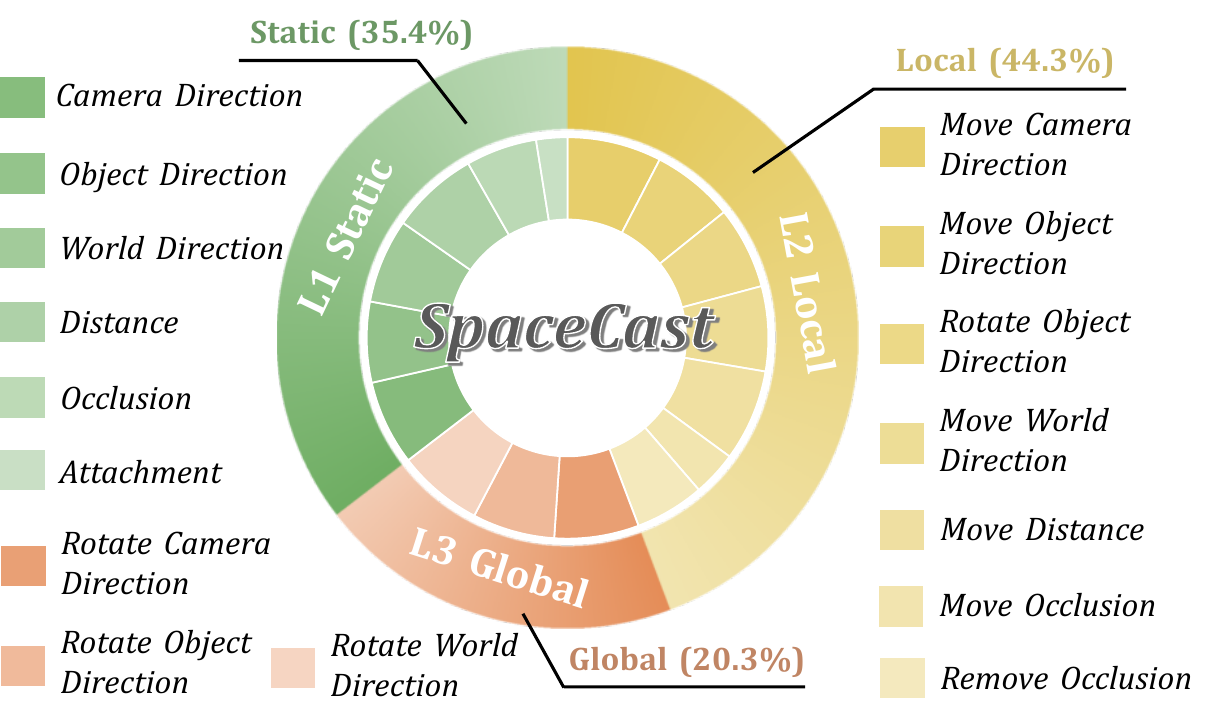}
    \vspace{-12pt}
    \caption{Task distribution of \dataname.}
    \label{fig:data-statistics}
    \vspace{-7pt}
\end{wrapfigure}

Figure~\ref{fig:data-statistics} summarizes the benchmark composition. The benchmark contains 3,862 questions across 182 scenes and 16 task types, drawn from the validation splits of ScanNet~\citep{dai2017scannet} and ScanNet++~\citep{yeshwanth2023scannet}. ScanNet contributes 718 questions from 134 scenes and ScanNet++ contributes 3,144 questions from 48 scenes. By capability level, 1,368 questions are L1 (Static Perception), 1,709 are L2 (Local Prediction), and 785 are L3 (Global Prediction). Among L2 and L3 questions, 2,217 change the queried relation after the transformation and 277 preserve it.

%% file: sections/experiments.tex
\section{Experiments}
\label{sec:experiments}

\subsection{Experimental Setup}
\label{sec:experimental-setup}

\paragraph{Models.}
We evaluate 21 models in 3 groups: 9 proprietary models, 6 general-purpose open-weight models, and 6 spatially specialized models. The proprietary set includes Claude Sonnet 4.6 and Claude Sonnet~5 \citep{anthropic2026claudesonnet46,anthropic2026claudesonnet5}; Gemini~3.5 Flash and Gemini~3 Flash \citep{google2026gemini35,google2025gemini3flash}; GPT-5.4 and GPT-5.6-Luna \citep{openai2026gpt54,openai2026gpt56}; Doubao 2.0 Pro \citep{bytedance2026seed20}; and Qwen3.6-Flash and Qwen3.7-Plus \citep{qwen2026visionmodels}. The general-purpose open-weight set comprises InternVL3.5-8B \citep{internvl2025internvl35}, Qwen3-VL-4B \citep{bai2025qwen3-vl}, and Qwen3.5 models at 2B, 4B, 9B, and 27B \citep{qwen2026qwen35}. The spatially specialized set contains VST-7B-RL \citep{yang2025vst}, SenseNova-SI-1.2 \citep{cai2026sensenova-si}, Cambrian-S-7B \citep{yang2025cambrians}, SpaceR \citep{ouyang2025spacer}, SpatialLadder \citep{li2026spatialladder}, and Spatial-MLLM-v1.1 \citep{wu2025spatial-mllm}. All models are evaluated with a chain-of-thought prompt and deterministic decoding.

\paragraph{Metrics.}
Our primary metric is the overall score, defined as the unweighted macro-average of per-task accuracy across all 16 task types. We additionally report Static, Local, and Global averages, each computed as the macro-average over the task types in the corresponding transformation family. See Appendix~\ref{app:experiments} for evaluation protocols, prompts, and more training details.

\input{tables/table_main-benchmark}

\input{tables/table_spatial-relation}

\subsection{Main Results}
\label{sec:main-results}

\paragraph{Overall Performance.}
Table~\ref{tab:main-benchmark} reports accuracy across all 16 task types. Human-level performance stands at 87.2\%, providing an upper bound against which all models fall substantially short. Among proprietary models, Gemini 3.5 Flash leads at 58.0\%, followed by Gemini 3 Flash (56.7\%) and Qwen3.7-Plus (55.3\%), with the remaining models ranging down to 36.0\% (Claude Sonnet 4.6); even the best model trails human level by 29.2 pp, highlighting the substantial headroom that remains. The strongest general-purpose open-weight model, Qwen3.5-27B, reaches 39.3\%, 18.7 pp below the top proprietary model and 47.9 pp below human level. Spatially specialized models score lowest overall (25.7--29.1\%), falling below every proprietary and general-purpose open-weight model despite being explicitly trained for spatial tasks, and remain near the random baseline of 25.9\%, suggesting that current spatial fine-tuning paradigms have yet to instill genuine spatial reasoning capabilities. See Appendix~\ref{app:case-study} for qualitative analysis and representative failure cases.

\paragraph{Capability Level Analysis.}
Predictive reasoning is broadly harder than static perception, though the pattern differs across model groups. Proprietary and general-purpose open-weight models both score lower on L2 than L1, with the largest drops for Gemini 3.5 Flash ($-21.8$ pp), Qwen3.7-Plus ($-20.6$ pp), and Qwen3.6-Flash ($-18.3$ pp). Spatially specialized models show a slightly higher group average on L2 than L1, yet all three groups decline on L3 relative to L1. The strongest proprietary models recover substantially at L3 (Gemini 3.5 Flash: L2 45.3\%, L3 69.4\%; Qwen3.7-Plus: L2 43.2\%, L3 66.6\%), reflecting strong performance on global rotation tasks where a holistic grasp of spatial layout appears sufficient. Spatially specialized models instead show markedly lower L3 averages (e.g., SenseNova-SI-1.2: L2 30.5\%, L3 16.1\%; Cambrian-S-7B: L2 28.8\%, L3 17.1\%), suggesting that spatial fine-tuning reinforces local perceptual features while leaving global spatial awareness undeveloped. These opposing patterns point to fundamentally different capability profiles rather than a uniform scaling of spatial reasoning.

\paragraph{Spatial Relation Analysis.}
Table~\ref{tab:spatial-relation} reports accuracy per spatial relation category, averaged over all task types within each category across L1--L3. Attachment yields the highest scores across selected proprietary and open-weight models (Qwen3.7-Plus: 80.4\%, GPT-5.6-Luna: 74.2\%, Qwen3.5-27B: 68.0\%), suggesting that binary contact judgements are comparatively tractable. Occlusion ranks second for most models, while direction and distance remain the most challenging relation types. Spatially specialized models score disproportionately low on attachment (SenseNova-SI-1.2: 13.4\%, SpatialLadder: 9.3\%), indicating that attachment relations are severely underrepresented in existing spatial fine-tuning data and task design.

\paragraph{Reference Frame Analysis.}
Table~\ref{tab:reference-frame} reports accuracy per directional reference frame, averaged over all direction-type tasks across L1--L3. World-centric reasoning is the strongest reference frame across both selected proprietary models (Qwen3.7-Plus: 58.4\%, GPT-5.6-Luna: 49.9\%), while camera-centric reasoning remains competitive for Qwen3.7-Plus (53.9\%) and GPT-5.6-Luna (45.7\%). Object-centric reasoning is consistently the weakest frame among proprietary and open-weight models, with Qwen3-VL-4B exhibiting the largest camera--object gap (39.3\% vs.\ 16.2\%). Spatially specialized models show no clear reference frame preference, with scores clustered in the 22--32\% range, suggesting a general deficit rather than frame-specific weaknesses.

\begin{keytakeaways}{Predictive Spatial Reasoning Remains Challenging}
\begin{itemize}[leftmargin=1.2em,itemsep=1pt,topsep=0pt,parsep=0pt]
    \item All evaluated models remain far below human spatial reasoning.
    \item General-purpose models outperform models explicitly fine-tuned for spatial tasks.
    \item Strong static perception does not reliably translate into spatial state updating.
\end{itemize}
\end{keytakeaways}

\begin{figure*}[t]
    \centering
    \begin{minipage}[t]{0.64\textwidth}
        \centering
        \includegraphics[width=\linewidth]{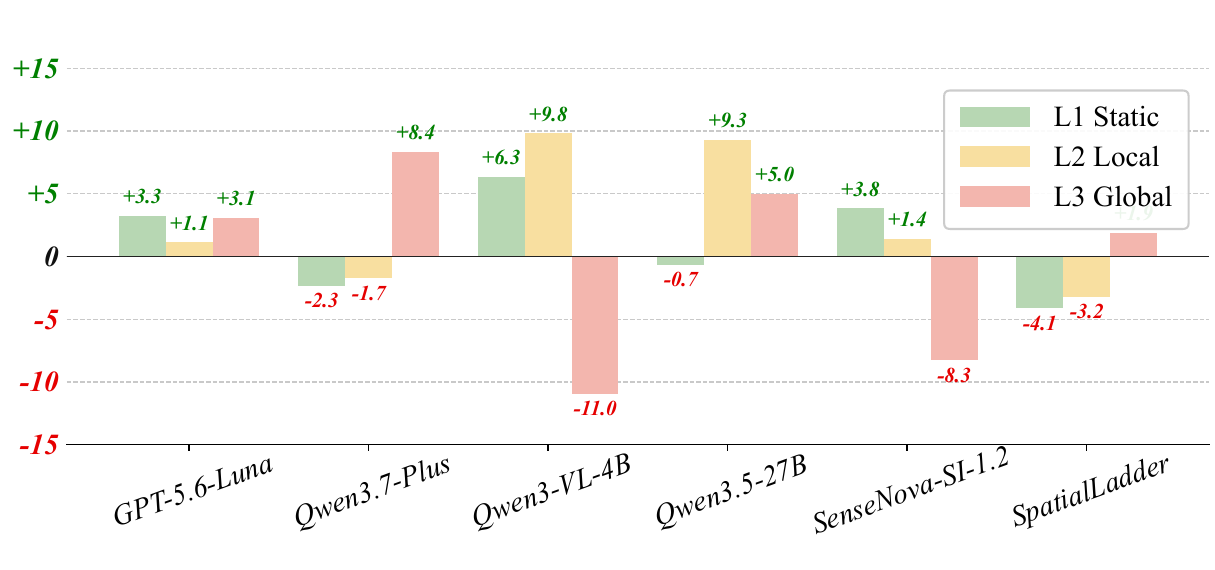}
        \vspace{-24pt}
        \captionof{figure}{CoT vs.\ Direct accuracy gap (pp) for six representative models, broken down by capability level.}
        \vspace{-10pt}
        \label{fig:cot-direct-level}
    \end{minipage}\hfill
    \begin{minipage}[t]{0.34\textwidth}
        \centering
        \includegraphics[width=\linewidth]{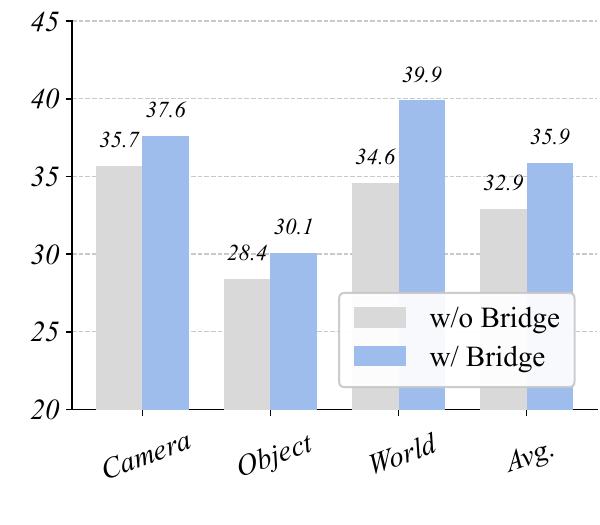}
        \vspace{-24pt}
        \captionof{figure}{Directional accuracy (\%) with and without bridge views.}
        \vspace{-10pt}
        \label{fig:bridge-ablation}
    \end{minipage}
\end{figure*}

\subsection{Further Analysis}

\paragraph{Effect of CoT Prompting.}
Figure~\ref{fig:cot-direct-level} shows the accuracy gap between CoT~\citep{wei2022cot} and Direct prompting for six representative models, broken down by capability level. CoT does not provide a uniform advantage: its effect varies sharply across model families and capability levels. It improves global tasks for Qwen3.7-Plus but harms its static and local performance, whereas Qwen3-VL-4B shows the opposite pattern. The two spatially specialized models are similarly mixed, indicating that explicit rationales interact with a model's existing spatial capability profile rather than reliably inducing predictive spatial reasoning.

\paragraph{Effect of Bridge Views.}
Figure~\ref{fig:bridge-ablation} shows the mean directional accuracy with and without bridge views across six representative models, averaged over three reference frames. Removing bridge views consistently lowers accuracy across all three reference frames. The largest drop occurs for world-centric direction ($39.9\%$ to $34.6\%$, $-5.3$ pp), while camera-centric ($-1.9$ pp) and object-centric ($-1.7$ pp) direction show smaller decreases. This pattern suggests that bridge views are particularly important for maintaining a scene-level spatial representation across views.

\paragraph{Effect of External Spatial Evidence.}
Recent work suggests that generative world models can support spatial reasoning by providing imagined observations of future scene states~\citep{zhang2026world2vlm,yang2026mindjourney,cao2026spatialdreamer}. Motivated by this, we test whether external evidence can alleviate different bottlenecks in our observe--transform--infer pipeline, using the same six representative models as in Figure~\ref{fig:bridge-ablation}. We consider two evidence families: explicit 3D evidence augments $S_0$ with a BEV image or textual 3D metadata, while world-model evidence approximates $S_1$ via post-intervention images from Qwen-Image-Edit~\citep{qwen2025qwenimageedit} or GPT-image-2~\citep{openai2026gptimage2}, or a transformation video from Seedance2-mini~\citep{bytedance2026seedance20mini}.

\begin{figure*}[t]
    \centering
    \includegraphics[width=\textwidth]{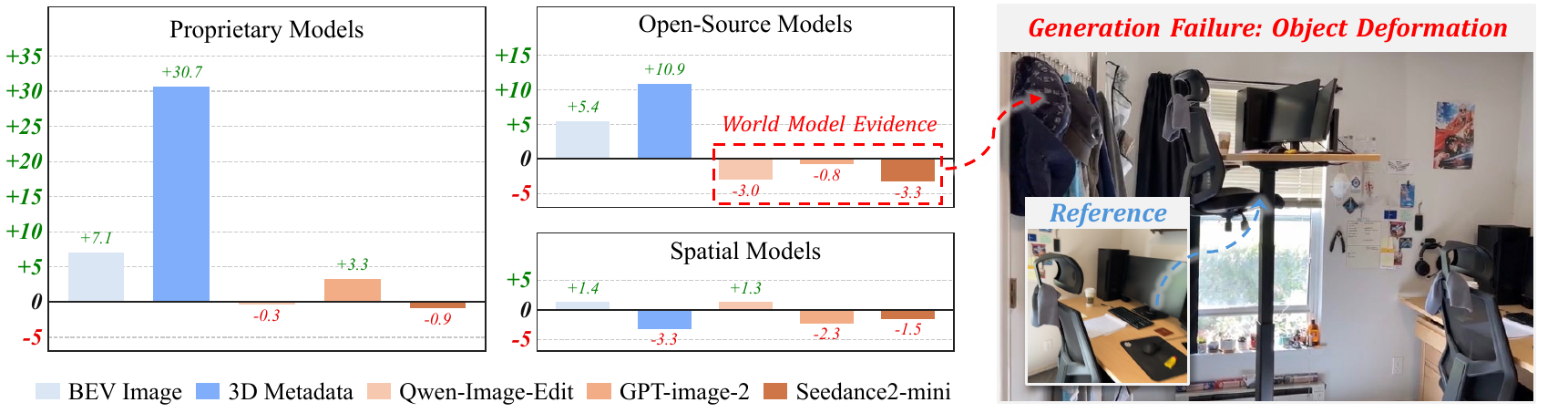}
    \caption{Effect of external spatial evidence. \textbf{Left:} Mean accuracy change (pp) relative to baseline across three model categories (proprietary, open-source, and spatial) when augmented with BEV images, textual 3D metadata, or generated outcome evidence from Qwen-Image-Edit, GPT-image-2, and Seedance2-mini. \textbf{Right:} A representative failure where the reference object is severely deformed in the generated outcome image.}
    \vspace{-10pt}
    \label{fig:external-spatial-evidence}
\end{figure*}

Figure~\ref{fig:external-spatial-evidence} shows that general-purpose models effectively exploit explicit 3D evidence. Both proprietary and open-weight models improve with BEV images, and textual 3D metadata yields substantially larger gains (+30.7 pp and +10.9 pp, respectively). The stronger response among proprietary models suggests that recovering 3D structure from visual observations is a primary bottleneck: once made explicit, these models can perform the subsequent transformation and relational reasoning. Spatially specialized models show little benefit and decline with 3D metadata ($-3.3$ pp), suggesting that fine-tuning without such evidence reduces their ability to exploit unfamiliar spatial representations.

World-model evidence produces much smaller and less consistent changes. To determine whether this reflects generation failures or ineffective evidence use, we manually audit generated endpoints for 600 local-intervention questions. Among incorrect responses across 6 representative models, 80.3\% under Qwen-Image-Edit and 38.3\% under GPT-image-2 are associated with generation failures, including artifacts, missing objects, deformation, and scale or perspective mismatch. Errors that persist after successful generation reveal a second bottleneck: even given an approximate $S_1$, models must still align the synthesized endpoint with the original multi-view evidence. Closing this gap thus requires advances on both generation fidelity and cross-view evidence integration.

\begin{keytakeaways}{Spatial Evidence Must Be Both Available and Usable}
\begin{itemize}[leftmargin=1.2em,itemsep=1pt,topsep=0pt,parsep=0pt]
    \item CoT prompting does not consistently improve predictive spatial reasoning.
    \item Bridge views are essential for integrating distributed observations into a global spatial state.
    \item Explicit 3D evidence is more reliable than generated outcomes.
\end{itemize}
\end{keytakeaways}

\input{tables/table_training-results}

\subsection{Training Experiments}

\paragraph{Training Setup.}
We train Qwen3-VL-4B with SFT and GRPO~\citep{shao2024deepseekmath} to assess whether standard post-training can improve predictive spatial reasoning and whether models trained on our dataset generalize to other spatial reasoning benchmarks. The training set is generated by the same programmatic pipeline, covering all 16 task types from scenes disjoint from the benchmark. We compare Mixed SFT (joint sampling over all task types) with Staged SFT (L1 first, then L2/L3 with L1 replay). SFT trains rank-32 LoRA~\citep{hu2021lora} adapters on all language-model linear layers and the multimodal aligner (vision encoder frozen), with peak learning rates of $10^{-4}$ and $10^{-5}$. GRPO uses the same staged schedule with rank-16 adapters and learning rates of $10^{-5}$ and $5\times10^{-6}$.

\paragraph{In-domain Results.}
Table~\ref{tab:training-results} shows that both SFT variants substantially improve predictive spatial reasoning on \dataname, with Mixed SFT reaching 64.8\% and Staged SFT reaching 65.7\% overall. Staged SFT improves Local performance from 51.3\% to 54.0\%, suggesting that progressively introducing higher-level task families benefits local state updating, while Mixed SFT is slightly stronger on Global tasks (73.1\% vs.\ 70.8\%). GRPO reaches 37.7\% overall and performs best on Global tasks (45.7\%), but remains substantially below both SFT variants, suggesting that task difficulty limits the model's ability to obtain consistent reward signals for effective on-policy training.

\paragraph{Out-of-domain Transfer.}
As Table~\ref{tab:training-results} shows, both SFT variants transfer beyond \dataname across six OOD benchmarks~\citep{yang2026mmsi-bench,mirzaee2021spartqa,wang2026mindcube,yi2020clevrer,yang2025thinking,zhang2025dsibench} spanning multi-image and language-grounded spatial reasoning, limited-view mental modeling, physical event understanding, and dynamic video-based spatial reasoning. Staged SFT achieves the strongest macro average (38.1\%), exceeding Mixed SFT (37.2\%) and the base model (33.5\%), with the largest gains on DSI-Bench (+10.2 pp), SPARTQA (+6.0 pp), CLEVRER (+5.8 pp), and VSI-Bench (+4.7 pp), while MindCube remains slightly below the base model. GRPO shows a distinct transfer profile, with substantial gains on MindCube and DSI-Bench, smaller improvements on VSI-Bench and MMSI, and slight declines on SPARTQA and CLEVRER. Despite their different transfer profiles, both SFT and GRPO improve the out-of-domain macro average over the base model, demonstrating that training on \dataname data yields transferable spatial reasoning gains regardless of the post-training paradigm.

\begin{keytakeaways}{Post-Training Improves Reasoning but Transfers Selectively}
\begin{itemize}[leftmargin=1.2em,itemsep=1pt,topsep=0pt,parsep=0pt]
    \item Supervised fine-tuning substantially improves in-domain predictive spatial reasoning.
    \item Staged training transfers more consistently than mixed training.
    \item Training on our data yields transferable spatial reasoning gains across both post-training paradigms.
\end{itemize}
\end{keytakeaways}

%% file: tables/table_main-benchmark.tex
\begin{table*}[t]
    \centering
    \scriptsize
    \definecolor{StaticShade}{HTML}{F5FDF3}
    \definecolor{LocalShade}{HTML}{FFFDE9}
    \definecolor{GlobalShade}{HTML}{FEF1E5}
    \definecolor{ModelGroupShade}{HTML}{F3FBFF}
    \setlength{\tabcolsep}{1.2pt}
    \newcolumntype{N}{>{\centering\arraybackslash}b{2em}}
    \newcommand{\modelgroupheader}[1]{%
        \rowcolor{ModelGroupShade}\textbf{\emph{#1}} & & &
        \multicolumn{19}{>{\columncolor{ModelGroupShade}}r}{} \\
    }
    \resizebox{\textwidth}{!}{%
    \begin{tabular}{r|cc|*{19}{N}}
        & &
        & \rotatebox{75}{Cam.\ Dir.} & \rotatebox{75}{Obj.\ Dir.} & \rotatebox{75}{World\ Dir.} & \rotatebox{75}{Dist.} & \rotatebox{75}{Occ.} & \rotatebox{75}{Att.} & \rotatebox{75}{\textbf{Avg.}}
        & \rotatebox{75}{Move\ Cam.\ Dir.} & \rotatebox{75}{Move\ Obj.\ Dir.} & \rotatebox{75}{Rotate\ Obj.\ Dir.} & \rotatebox{75}{Move\ World\ Dir.} & \rotatebox{75}{Move\ Dist.} & \rotatebox{75}{Move\ Occ.} & \rotatebox{75}{Remove\ Occ.} & \rotatebox{75}{\textbf{Avg.}}
        & \rotatebox{75}{Rotate\ Cam.\ Dir.} & \rotatebox{75}{Rotate\ Obj.\ Dir.} & \rotatebox{75}{Rotate\ World\ Dir.} & \rotatebox{75}{\textbf{Avg.}} \\
        \textbf{Model} & \textbf{Rank} & \textbf{Overall}
        & \multicolumn{7}{c}{\cellcolor{StaticShade}\textbf{L1: Static}}
        & \multicolumn{8}{c}{\cellcolor{LocalShade}\textbf{L2: Local}}
        & \multicolumn{4}{c}{\cellcolor{GlobalShade}\textbf{L3: Global}} \\
        \hline
        \modelgroupheader{Baseline}
        Human Level & -- & 87.2 & 92.0 & 92.4 & 90.1 & 89.7 & 100.0 & 97.9 & 93.7 & 81.6 & 83.2 & 83.6 & 76.1 & 82.5 & 84.2 & 97.2 & 84.1 & 81.0 & 82.4 & 80.8 & 81.4 \\
        Random Level & -- & 25.9 & 25.8 & 25.9 & 24.0 & 25.7 & 29.9 & 11.3 & 23.8 & 23.5 & 21.1 & 24.2 & 23.1 & 21.8 & 37.4 & 38.0 & 27.0 & 24.3 & 31.6 & 26.3 & 27.4 \\
        \hline
        \modelgroupheader{Proprietary Models}
        Claude Sonnet 4.6 & 9 & 36.0 & 29.5 & 17.1 & 35.0 & 40.1 & 71.5 & 66.0 & 43.2 & 38.2 & 14.5 & 30.1 & 30.3 & 30.5 & 43.9 & 40.7 & 32.6 & 39.9 & 15.6 & 33.8 & 29.8 \\
        Claude Sonnet 5 & 7 & 38.3 & 31.1 & 26.3 & 38.4 & 32.4 & 50.7 & 56.7 & 39.2 & 34.1 & 29.3 & 30.5 & 35.2 & 40.4 & 50.4 & 27.8 & 35.4 & 41.1 & 50.4 & 38.7 & 43.4 \\
        Gemini 3.5 Flash & \textbf{1} & \textbf{58.0} & \textbf{62.9} & 38.6 & \textbf{71.9} & \textbf{70.2} & 78.3 & 80.4 & \textbf{67.1} & \textbf{50.9} & 41.8 & 34.4 & 50.8 & 35.8 & 34.5 & 69.0 & 45.3 & \textbf{63.5} & 76.6 & \textbf{68.0} & \textbf{69.4} \\
        Gemini 3 Flash & 2 & 56.7 & 46.6 & 33.5 & 64.6 & 61.4 & 80.5 & \textbf{82.5} & 61.5 & 45.4 & \textbf{52.7} & \textbf{37.1} & 52.7 & 37.2 & 43.2 & \textbf{75.5} & \textbf{49.1} & 51.3 & 83.2 & 60.2 & 64.9 \\
        GPT-5.4 & 8 & 37.7 & 32.2 & 18.7 & 46.4 & 43.8 & 60.6 & 49.5 & 41.9 & 39.2 & 17.6 & 21.1 & 43.2 & \textbf{43.9} & \textbf{56.1} & 44.0 & 37.9 & 31.9 & 21.5 & 33.1 & 28.8 \\
        GPT-5.6-Luna & 5 & 46.6 & 45.8 & 27.9 & 47.9 & 47.8 & 71.0 & 74.2 & 52.4 & 45.7 & 34.4 & 26.2 & 53.0 & 34.0 & 31.7 & 42.1 & 38.2 & 45.6 & 69.1 & 48.9 & 54.5 \\
        Doubao 2.0 Pro & 6 & 45.6 & 38.6 & 18.7 & 63.9 & 32.4 & \textbf{85.1} & 69.1 & 51.3 & 30.7 & 44.5 & 30.1 & 44.7 & 30.5 & 46.8 & 46.8 & 39.2 & 47.5 & 41.0 & 59.0 & 49.2 \\
        Qwen3.6-Flash & 4 & 47.7 & 44.3 & 34.3 & 57.4 & 55.5 & 65.2 & 72.2 & 54.8 & 42.0 & 21.5 & 30.5 & 47.0 & 35.1 & 40.3 & 39.4 & 36.5 & 54.4 & 76.2 & 47.7 & 59.4 \\
        Qwen3.7-Plus & 3 & 55.3 & 50.8 & 44.2 & 68.1 & 59.9 & 79.2 & 80.4 & 63.8 & 49.8 & 34.4 & 27.7 & \textbf{55.3} & 42.8 & 41.0 & 51.4 & 43.2 & 61.2 & \textbf{86.7} & 51.9 & 66.6 \\
        \hline
        \modelgroupheader{Open-Source Models}
        InternVL3.5-8B & 6 & 30.4 & 36.7 & 11.6 & 31.2 & 30.1 & 52.5 & 35.1 & 32.9 & 28.0 & 25.4 & 23.0 & 26.1 & 40.4 & 44.6 & 23.6 & 30.2 & 29.7 & 9.4 & 39.5 & 26.2 \\
        Qwen3-VL-4B & 3 & 34.0 & 49.2 & 6.0 & 35.4 & 46.0 & 60.6 & 42.3 & 39.9 & 36.9 & 28.5 & 24.2 & 29.5 & 31.9 & 51.8 & 32.9 & 33.7 & 31.9 & 6.3 & 30.1 & 22.8 \\
        Qwen3.5-2B & 5 & 31.3 & 41.7 & 25.9 & 27.0 & 27.2 & 46.2 & 39.2 & 34.5 & 34.5 & 18.8 & 28.1 & 32.6 & 27.0 & 37.4 & 40.7 & 31.3 & 27.4 & 14.5 & 32.7 & 24.8 \\
        Qwen3.5-4B & 4 & 33.7 & 36.4 & 17.1 & 31.9 & 38.6 & 71.0 & 60.8 & 42.7 & 34.5 & 23.8 & 20.7 & 31.8 & 22.8 & 43.2 & 38.0 & 30.7 & 30.4 & 11.3 & 26.7 & 22.8 \\
        Qwen3.5-9B & 2 & 34.6 & 40.5 & 10.8 & 33.5 & 34.6 & 70.6 & 48.5 & 39.7 & 29.7 & 21.9 & 26.2 & 32.2 & 40.0 & 49.6 & 36.1 & 33.7 & 34.6 & 14.1 & 31.6 & 26.7 \\
        Qwen3.5-27B & \textbf{1} & 39.3 & 34.8 & 22.7 & 41.4 & 40.8 & 66.1 & 68.0 & 45.7 & 40.3 & 18.0 & 25.8 & 37.5 & 31.6 & 51.1 & 43.5 & 35.4 & 40.3 & 29.3 & 37.2 & 35.6 \\
        \hline
        \modelgroupheader{Spatial Models}
        VST-7B-RL & 6 & 25.7 & 14.8 & 32.7 & 34.2 & 25.4 & 31.7 & 15.5 & 25.7 & 25.6 & 32.4 & 28.1 & 28.8 & 32.3 & 30.9 & 24.5 & 29.0 & 15.6 & 12.5 & 26.3 & 18.1 \\
        SenseNova-SI-1.2 & \textbf{1} & 29.1 & 15.2 & \textbf{62.2} & 35.0 & 30.1 & 47.5 & 13.4 & 33.9 & 39.9 & 27.3 & 23.4 & 30.7 & 27.0 & 33.8 & 31.5 & 30.5 & 21.7 & 4.3 & 22.2 & 16.1 \\
        Cambrian-S-7B & 3 & 27.8 & 15.2 & 41.0 & 22.8 & 34.2 & 51.1 & 27.8 & 32.0 & 29.0 & 29.3 & 31.3 & 25.0 & 23.5 & 52.5 & 11.1 & 28.8 & 14.8 & 9.0 & 27.4 & 17.1 \\
        SpaceR & 4 & 27.7 & 32.6 & 10.8 & 27.0 & 26.1 & 39.4 & 39.2 & 29.2 & 30.7 & 19.5 & 22.7 & 15.5 & 28.1 & 44.6 & 25.9 & 26.7 & 26.6 & 22.3 & 31.6 & 26.8 \\
        SpatialLadder & 2 & 27.9 & 18.2 & 20.3 & 25.5 & 33.1 & 43.4 & 9.3 & 25.0 & 38.6 & 32.0 & 33.2 & 36.7 & 28.8 & 47.5 & 28.2 & 35.0 & 11.0 & 8.6 & 32.7 & 17.4 \\
        Spatial-MLLM-v1.1 & 5 & 26.4 & 4.5 & 56.2 & 22.1 & 31.3 & 33.5 & 0.0 & 24.6 & 43.3 & 43.8 & 23.0 & 38.3 & 22.5 & 24.5 & 33.3 & 32.7 & 8.4 & 2.3 & 35.3 & 15.3 \\
    \end{tabular}%
    }
    \vspace{-6pt}
    \caption{Accuracy (\%) on all 16 \dataname task types. ``Cam.'', ``Obj.'', and ``World'' denote camera-centric, object-centric, and world-centric reference frames; ``Dir.'', ``Dist.'', ``Occ.'', and ``Att.'' denote direction, distance, visibility, and attachment. ``Move'', ``Rotate'', and ``Remove'' denote local intervention types. \textbf{Bold} indicates the best model result in each column; baselines are excluded.}
    \label{tab:main-benchmark}
\end{table*}

%% file: tables/table_spatial-relation.tex
\begin{table*}[t]
    \centering
    \scriptsize
    \definecolor{AnalysisModelShade}{HTML}{F3FBFF}
    \definecolor{AnalysisRelationShade}{HTML}{FFFDE9}
    \definecolor{AnalysisFrameShade}{HTML}{FEF1E5}
    \renewcommand{\arraystretch}{1.05}
    \setlength{\tabcolsep}{2.5pt}
    \begin{minipage}[t]{0.55\textwidth}
        \centering
        \begin{tabularx}{\linewidth}[t]{l|*{4}{>{\centering\arraybackslash}X}}
            \toprule
            \rowcolor{AnalysisRelationShade}\cellcolor{AnalysisModelShade}\textbf{Model} & \textbf{Direction} & \textbf{Distance} & \textbf{Occlusion} & \textbf{Attachment} \\
            \midrule
            GPT-5.6-Luna & 44.5 & 40.9 & 48.3 & 74.2 \\
            Qwen3.7-Plus & \textbf{53.0} & \textbf{51.4} & \textbf{57.2} & \textbf{80.4} \\
            \midrule
            Qwen3-VL-4B & 27.8 & 39.0 & 48.4 & 42.3 \\
            Qwen3.5-27B & 32.7 & 36.2 & 53.6 & 68.0 \\
            \midrule
            SenseNova-SI-1.2 & 28.2 & 28.6 & 37.6 & 13.4 \\
            SpatialLadder & 25.7 & 31.0 & 39.7 & 9.3 \\
            \bottomrule
        \end{tabularx}
        \captionof{table}{Accuracy (\%) per spatial relation category, averaged over all task types across L1--L3. \textbf{Bold} indicates the highest result in each column.}
        \vspace{-12pt}
        \label{tab:spatial-relation}
    \end{minipage}\hspace{0.02\textwidth}%
    \begin{minipage}[t]{0.41\textwidth}
        \centering
        \begin{tabularx}{\linewidth}[t]{*{3}{>{\centering\arraybackslash}X}}
            \toprule
            \rowcolor{AnalysisFrameShade}\textbf{Camera-Centric} & \mbox{\textbf{Object-Centric}} & \textbf{World-Centric} \\
            \midrule
            45.7 & 39.4 & 49.9 \\
            \textbf{53.9} & \textbf{48.3} & \textbf{58.4} \\
            \midrule
            39.3 & 16.2 & 31.7 \\
            38.5 & 24.0 & 38.7 \\
            \midrule
            25.6 & 29.3 & 29.3 \\
            22.6 & 23.5 & 31.6 \\
            \bottomrule
        \end{tabularx}
        \captionof{table}{Accuracy (\%) per directional reference frame. \textbf{Bold} indicates the highest result in each column.}
        \vspace{-12pt}
        \label{tab:reference-frame}
    \end{minipage}
\end{table*}

%% file: tables/table_training-results.tex
\begin{table}[t]
    \centering
    \scriptsize
    \setlength{\tabcolsep}{3.0pt}
    \definecolor{ModelGroupShade}{HTML}{F3FBFF}
    \definecolor{GainGreen}{HTML}{5B8C5A}
    \definecolor{LossRed}{HTML}{E84537}
    \newcommand{\gain}[1]{\raisebox{-0.55ex}{\textit{\tiny\textcolor{GainGreen}{\ensuremath{\uparrow}#1}}}}
    \newcommand{\loss}[1]{\raisebox{-0.55ex}{\textit{\tiny\textcolor{LossRed}{\ensuremath{\downarrow}#1}}}}
    \begin{adjustbox}{width=\textwidth}
    \begin{tabular}{l*{11}{c}}
        \toprule
        \multirow{2}{*}{\textbf{Method}}
        & \multicolumn{4}{c}{\textbf{\dataname}}
        & \multicolumn{7}{c}{\textbf{OOD Benchmarks}} \\
        \cmidrule(lr){2-5}\cmidrule(lr){6-12}
        & \textbf{Static} & \textbf{Local} & \textbf{Global} & \textbf{Overall}
        & \textbf{MMSI} & \textbf{SPARTQA} & \textbf{MindCube} & \textbf{CLEVRER} & \textbf{VSI} & \textbf{DSI} & \textbf{Avg.} \\
        \midrule
        \rowcolor{ModelGroupShade}
        Qwen3-VL-4B
        & 39.9 & 33.7 & 22.8 & 34.0
        & 27.4 & 28.8 & 31.2 & 35.7 & 40.7 & 37.3 & 33.5 \\
        \hspace{0.25em}\textit{+ Mixed SFT}
        & 76.4\gain{36.5} & 51.3\gain{17.6} & 73.1\gain{50.3} & 64.8\gain{30.8}
        & 28.8\gain{1.5} & 33.6\gain{4.8} & 29.1\loss{2.1} & 41.8\gain{6.2} & 44.7\gain{4.0} & 45.2\gain{7.9} & 37.2\gain{3.7} \\
        \hspace{0.25em}\textit{+ Staged SFT}
        & 76.7\gain{36.8} & 54.0\gain{20.3} & 70.8\gain{48.0} & 65.7\gain{31.7}
        & 28.5\gain{1.1} & 34.8\gain{6.0} & 30.8\loss{0.3} & 41.5\gain{5.8} & 45.3\gain{4.7} & 47.4\gain{10.2} & 38.1\gain{4.6} \\
        \hspace{0.25em}\textit{+ GRPO}
        & 40.3\gain{0.4} & 32.0\loss{1.7} & 45.7\gain{22.9} & 37.7\gain{3.7}
        & 27.7\gain{0.3} & 26.4\loss{2.4} & 45.2\gain{14.0} & 35.2\loss{0.5} & 44.0\gain{3.3} & 47.8\gain{10.5} & 37.7\gain{4.2} \\
        \bottomrule
    \end{tabular}
    \end{adjustbox}
    \caption{Accuracy (\%) of Qwen3-VL-4B before and after Mixed SFT, Staged SFT, and GRPO on \dataname and six OOD spatial reasoning benchmarks. MMSI, VSI, and DSI denote MMSI-Bench, VSI-Bench, and DSI-Bench, respectively.}
    \label{tab:training-results}
\end{table}

%% file: sections/conclusion.tex
\section{Conclusion}

We introduced \dataname{}, the first benchmark to directly and diagnostically evaluate predictive spatial reasoning, where tasks progress from understanding observed scenes to updating spatial states and inferring relations in unobserved outcomes. Experiments reveal a substantial human--model gap and show that existing spatial training does not transfer to spatial state updating. Fine-tuning on our programmatically generated data yields consistent gains both in-domain and across six out-of-domain benchmarks, suggesting that predictive spatial reasoning represents a distinct and largely unsolved capability that \dataname{} is positioned to drive progress on.

%% file: sections/statement.tex
\subsection*{AI Use Statement}
Generative AI tools were used to assist with code implementation, figure plotting, and manuscript polishing. All AI-assisted content was reviewed and verified by the authors, who take full responsibility for the final work.

\subsection*{Ethics Statement}
This work uses the publicly available ScanNet and ScanNet++ datasets in accordance with their respective terms of use. Human involvement was limited to benchmark verification and baseline evaluation, and no sensitive personal information was collected. The benchmark studies hypothetical spatial transformations and does not operate physical systems. Nevertheless, the evaluated models remain unreliable, and their outputs should not be interpreted as evidence of readiness for safety-critical embodied deployment.

\subsection*{Reproducibility Statement}
The benchmark and experimental pipeline are documented in detail to support reproducibility. Appendix~\ref{app:benchmark-details} specifies the data sources and splits, task definitions, view-chain construction, spatial relations and transformations, question generation, and quality-control procedures. Appendix~\ref{app:experiments} covers the evaluation protocol and baselines, outcome-media generation, training data and hyperparameters, out-of-domain evaluation subsets, and complete evaluation and world-model prompts.

%% file: sections/appendix.tex
\newpage

\appendix

\section{Limitations}
\label{app:limitations}

\dataname comprises 3,862 questions drawn from 182 ScanNet and ScanNet++ scenes, all of which are indoor environments; ScanNet++ alone contributes 3,144 questions. Consequently, the benchmark does not cover outdoor settings or object arrangements and scales absent from these two sources. Question labels depend on reconstructed meshes, depth maps, camera poses, and instance-level semantic annotations, so geometric errors can affect ray-cast visibility, apparent surface contact or support relationships, and whether a queried object is sufficiently recognizable in a selected view. We exclude geometrically ambiguous cases and require independent review by at least two annotators per question, though these measures cannot fully eliminate reconstruction-related uncertainty in occlusion and attachment labels, particularly for objects with incomplete geometry or low reconstruction confidence. Finally, the three intervention types (translations, rotations, and removals) update scene geometry programmatically without simulating physical dynamics, so the benchmark does not assess deformation, friction, or collision-driven motion effects.

\section{\dataname Details}
\label{app:benchmark-details}

\subsection{Data Sources and Scene Splits}
\label{app:data-sources}

\dataname is generated from the validation splits of ScanNet and ScanNet++, which provide registered RGB-D observations, calibrated camera poses, reconstructed scene geometry, and instance-level semantic annotations. The evaluation set comprises 3,862 unique questions spanning 16 task types across 182 indoor scenes: 718 questions from 134 ScanNet scenes and 3,144 from 48 ScanNet++ scenes. The same generation pipeline independently produces a training pool of 12,249 unique questions covering all 16 task types from 54 disjoint training scenes. No scene identifier is shared between the training pool and the evaluation set.

\input{tables/table_app-task-composition}

\subsection{Task Definitions and Dataset Composition}
\label{app:task-definitions}

Table~\ref{tab:app-task-composition} lists the complete task suite by capability level, queried relation, and reference frame. The L1/L2/L3 levels describe the scope of spatial state change rather than a single scalar difficulty scale.

Of the evaluation questions, 1,368 belong to L1 (Static Perception), 1,709 to L2 (Local Prediction), and 785 to L3 (Global Prediction). The 2,494 L2 and L3 questions include 2,217 whose queried relation changes under the specified transformation and 277 whose relation remains unchanged, supporting evaluation of both spatial updating and invariance.

\begin{wrapfigure}[14]{r}{0.45\textwidth}
    \vspace{-8pt}
    \centering
    \includegraphics[width=\linewidth]{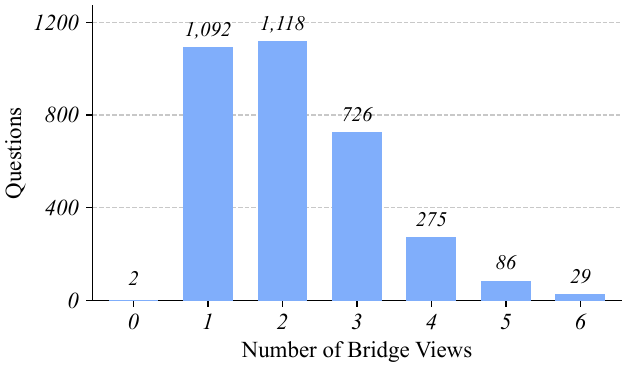}
    \vspace{-16pt}
    \caption{Distribution of retained bridge views across the 3,328 cross-view questions.}
    \label{fig:app-bridge-count}
    \vspace{-12pt}
\end{wrapfigure}

\subsection{Anchor--Bridge View Construction}
\label{app:anchor-bridge}

For each cross-view question, the generator partitions the queried object roles into two identity-disjoint groups, $\mathcal{G}_A$ and $\mathcal{G}_B$. Candidate anchor views are filtered by camera validity, image quality, projected object area, and geometric visibility. Anchor $A$ is selected only when every role in $\mathcal{G}_A$ is referable in $A$ and no role in $\mathcal{G}_B$ is meaningfully visible there; anchor $B$ satisfies the symmetric condition. This target disjointness applies to the queried objects, not to the entire image content.

The generator then connects the aggregate 3D centers $c_A$ and $c_B$ of the two groups with a normalized object-space corridor,
\begin{equation}
    p(t)=c_A+t(c_B-c_A),\qquad t\in[0,1].
\end{equation}
The segment is sampled at approximately 0.09\,m spacing, with 10--40 samples depending on its length. Camera calibration projects these samples into every candidate frame. A projected sample contributes to frame coverage only if it lies inside the image and the observed sensor depth is consistent with the expected camera-to-sample distance. Consecutive accepted samples form one or more normalized coverage intervals.

A valid transition between consecutive views must overlap the current coverage frontier by at least $\tau=0.10$ of the normalized corridor and advance toward the opposite anchor. Candidate bridge views that meaningfully expose both queried groups are removed. A directed graph search then identifies a low-cost complete route, and redundant views are pruned while coverage constraints are rechecked. The final model input is ordered as $A, V_1, \ldots, V_k, B$, where $V_1, \ldots, V_k$ are the numbered bridge views and $k \leq 6$. All retained views show only the pre-transformation scene, leaving the result of any intervention unobserved. The 10\% threshold refers to overlap between corridor-coverage intervals, not pixel intersection-over-union or a percentage of shared visual content.

Thirteen task types use this disjoint-anchor construction; the remaining three require only a single anchor view. Figure~\ref{fig:app-bridge-count} reports the number of retained bridge views across the 3,328 cross-view questions. Two questions connect the target-disjoint anchors without any intermediate bridge because the anchor coverage intervals already provide a valid handoff; all other cross-view questions contain at least one bridge view.

\subsection{Spatial Relations and 3D Transformations}
\label{app:geometry}

\paragraph{Direction.}
Direction is computed from the relative vector between object centers. Its floor-plane projection is quantized into eight 45-degree sectors, while a separate vertical rule assigns \texttt{above} or \texttt{below}. Camera-centric questions express the vector in the designated camera's horizontal axes, object-centric questions use a local frame defined by the reference object's stated facing direction, and world-centric questions use the scene's cardinal axes. Pairs near a directional decision boundary are excluded.

\paragraph{Distance.}
Distance approximates the shortest surface-to-surface separation between two instances using sampled mesh points with local triangle refinement. When surface samples are unavailable, the fallback uses the closest points on their 3D bounding boxes. The result is mapped to four categories: very close ($<1.0$\,m), close ($1.0$--$2.0$\,m), moderate ($2.0$--$3.3$\,m), and far ($>3.3$\,m). Questions within 0.1\,m of a category boundary are excluded.

\paragraph{Occlusion.}
The relation engine samples points on a target mesh, projects them into the calibrated camera, and ray casts each sample against the reconstructed scene mesh. A sample is considered visible when the target is the first valid intersection at the expected distance; a closer non-target intersection indicates external occlusion. Aggregated ray outcomes and camera-facing probes together distinguish three states: not occluded, occluded, and not visible. Removal questions recompute visibility after the specified deletion, while movement questions compute directed pairwise occlusion between the queried objects.

\paragraph{Attachment.}
The generator constructs a directed attachment graph from 3D contact, horizontal support overlap or containment, and semantic compatibility. Every child has at most one parent, and cycle-forming edges are rejected. The transitive descendants of a moved root define the component that moves with it; attachment-chain questions require exact-set identification of these descendants.

\paragraph{Transformations.}
L1 evaluates relations in the unmodified scene. For L2, object translations displace a selected root and its attachment descendants in the specified movement frame; object rotations move the component along a horizontal orbit of 45, 90, 135, or 180 degrees around a static pivot while preserving the moved object's intrinsic facing direction. Removal applies the specified deletion before visibility is recomputed. For L3, global transformations rotate the complete layout by 90, 180, or 270 degrees while keeping the designated camera fixed. The relation engine is rerun on the resulting 3D state to obtain the programmatic label. Candidates that would leave the estimated room bounds or introduce a new strict 3D bounding-box collision are rejected.

\subsection{Question Generation}
\label{app:question-generation}

Question templates instantiate the transformation, object roles, requested reference frame, and answer candidates. Template variants for L1, L2, and L3 are shown in Tables~\ref{tab:app-question-templates-l1}--\ref{tab:app-question-templates-l3}. Direction and distance tasks use four answer choices. Horizontal direction candidates are spaced at 90-degree intervals on the eight-direction ring so that adjacent directional labels do not serve as ambiguous distractors. Occlusion tasks use three choices, while attachment-chain questions are four-option multiple-selection items scored by exact set match. Semantic ground truth is computed before option order is randomized, and answer positions are balanced across the generated pool.

\input{tables/table_app-question-templates-l1}
\input{tables/table_app-question-templates-l2}
\input{tables/table_app-question-templates-l3}

\subsection{Quality Control}
\label{app:question-verification}

Automatic filtering removes geometrically ambiguous relations, distance labels near category boundaries, invalid motion endpoints, insufficiently grounded objects, and near-duplicate questions. Because finite 3D reconstruction accuracy is especially consequential for occlusion, attachment, and referability, these properties also undergo human verification. Ten annotators reviewed the benchmark; every question was independently assessed by at least two annotators who were blind to each other's decisions and to the programmatic answer. Disagreements were referred to a third annotator, who first made a blind judgment from the model-visible input and could consult geometry and generation traces only afterward.

Prior to adjudication, the 3,801 single-answer questions received 7,602 initial ratings with 98.9\% pairwise answer agreement. Reviewers disagreed on the answer for 42 questions and on answerability for a further 42 questions; the two categories overlapped on 15 questions, yielding 69 unique questions sent to third-review adjudication. For the 97 attachment-chain questions, reviewers achieved 91.8\% exact-set agreement. The complete quality-control process produced 157 ground-truth corrections, 143 question or option rewrites, and 36 removals, with 38 questions receiving both a ground-truth correction and a rewrite. All rewritten questions were independently reviewed by two annotators before inclusion.

\newpage

\section{Experimental Details}
\label{app:experiments}

\subsection{Evaluation Protocol and Metrics}
\label{app:evaluation-details}

The primary \dataname evaluation uses a standardized chain-of-thought (CoT) prompt with deterministic decoding. The CoT condition requests brief reasoning followed by the final option on the last line; the Direct condition requests only the final option. Single-choice questions are scored by the selected option, and attachment-chain questions require an exact match to the canonical option set. Paired diagnostics compare identical question identifiers under different prompting or evidence conditions.

The primary Overall score is the macro-average of the 16 task-type accuracies, with each task type weighted equally. Static, Local, and Global scores average the task-type accuracies within each capability level; relation-type and reference-frame summaries are computed analogously.

\subsection{Human and Random Baselines}
\label{app:human-random-baselines}

Twelve volunteers answered the benchmark using only the model-visible views and answer options, without additional guidance. This evaluation was conducted independently from the ten-annotator quality-control review. Their type-macro accuracy was 87.2\%, with 3,340 correct answers out of 3,862 questions. For the random baseline, we use seed 42 to sample uniformly from the displayed options for single-choice questions and from the eight valid answer sets for attachment-chain questions.

\subsection{Outcome Media Generation}
\label{app:outcome-media}

For each local-intervention question, we select an original RGB view covering the object's post-intervention location and use depth and camera geometry to verify its visibility. The post-intervention 3D location is projected into the view to obtain a target mask. Qwen-Image-Edit and GPT-image-2 each receive the original RGB view, the intervention description, and the mask to generate an endpoint image. Answering models then receive only the generated image as additional evidence.

Seedance2-mini receives the original anchor image, the GPT-image-2 endpoint image, and the intervention specification to generate a motion video. All answering models receive the same generated media for each question and condition.

\begin{figure*}[t]
    \centering
    \includegraphics[width=\textwidth]{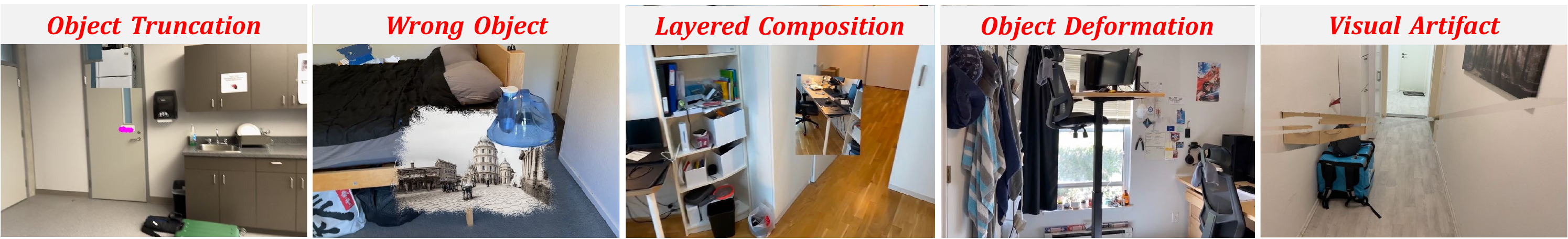}
    \caption{Representative endpoint-image generation failures. From left to right: object truncation, incorrect object identity, layered composition, object deformation, and visual artifacts.}
    \vspace{-8pt}
    \label{fig:wm-failure-cases}
\end{figure*}

Figure~\ref{fig:wm-failure-cases} shows representative endpoint-image failures identified during the manual audit.

\subsection{Training Setup}
\label{app:training}

\begin{wrapfigure}[11]{r}{0.23\textwidth}
    \vspace{-15pt}
    \centering
    \includegraphics[width=\linewidth]{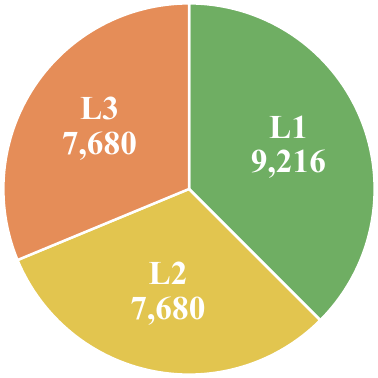}
    \vspace{-12pt}
    \caption{Training data composition.}
    \label{fig:training-data-composition}
    \vspace{-7pt}
\end{wrapfigure}

\paragraph{Training Data.}
We train Qwen3-VL-4B on 12,249 unique questions spanning all 16 task types from 54 and ScanNet++ scenes disjoint from the benchmark. Task-balanced repetitions yield 24,576 training exposures, whose composition by capability level is illustrated in Figure~\ref{fig:training-data-composition}.

\paragraph{Hyperparameters.}
SFT trains rank-32 LoRA adapters (scaling factor 64, dropout 0.05) on all language-model linear layers and trains the multimodal aligner, with the vision encoder frozen. It uses bfloat16 precision, AdamW with cosine decay, peak learning rates of $10^{-4}$ for the language adapter and $10^{-5}$ for the aligner, a global batch size of 32, a warm-up ratio of 0.03, and a maximum sequence length of 8,192.

GRPO trains rank-16 language-model LoRA adapters (scaling factor 32, dropout 0.05), with the vision encoder and multimodal aligner frozen. It uses groups of eight completions and a reward combining exact-answer correctness with a 0.1-weighted format term. Stage-1 and Stage-2 learning rates are $10^{-5}$ and $5\times10^{-6}$, respectively; the KL coefficient is $10^{-3}$, the clipping parameter is 0.2, the temperature is 1.0, and top-$p$ is 1.0.

\subsection{Evaluation of Trained Models}

\input{tables/table_sft-grpo-base-detailed}

\paragraph{In-Domain Evaluation.}
\label{app:sft-grpo-detailed}
Table~\ref{tab:sft-grpo-base-detailed} reports detailed per-task and capability-level results for Qwen3-VL-4B after Mixed SFT, Staged SFT, and GRPO, alongside the base-model results from Table~\ref{tab:main-benchmark}.

\paragraph{Out-of-Domain Evaluation.}
\label{app:ood}
Table~\ref{tab:app-ood-types} lists the question-type subsets and total counts used for the six OOD benchmarks. SPARTQA includes 50 questions from each of five listed types. VSI includes 50 questions from each of three listed types, sampled with seed 42; its relative-direction subset contains 30 easy, 5 medium, and 15 hard questions. All models are evaluated on the same selected questions.

\input{tables/table_app-ood-types}

\subsection{Evaluation Prompt}
\label{app:evaluation-prompt}

The main evaluation uses the following CoT prompts. The question placeholder contains the question text and its answer options.

\begin{AIbox}{System Prompt}
You are a careful vision-language assistant solving multiple-choice spatial reasoning questions about images.
\end{AIbox}

\begin{AIbox}{User Prompt: Single-Choice Questions}
\promptslot{question}\\
Work through your reasoning briefly, then give your final choice as the LAST line. Keep your reasoning as short as possible (a few sentences at most) so the final Answer line is not truncated. Return this format:\\
Reasoning: \promptangle{brief reasoning}\\
Answer: \promptangle{single letter}
\end{AIbox}

\begin{AIbox}{User Prompt: Attachment-Chain Questions}
\promptslot{question}\\
Work through your reasoning briefly, then give your final choice as the LAST line. If more than one option is correct, list all letters comma-separated. Keep your reasoning short so the final Answer line is not truncated. Return this format:\\
Reasoning: \promptangle{brief reasoning}\\
Answer: \promptangle{letter(s)}
\end{AIbox}

For the BEV condition, a bird's-eye image of the initial layout is appended after the original perspective views; the same answer format is used.

\raggedbottom
\subsection{World-Model Generation Prompts}
\label{app:world-model-prompts}

The endpoint-image pipeline inserts the moving group at its target location; Qwen-Image-Edit additionally removes the group from its original location and restores the vacated background. The two-panel input places the group's appearance reference on the left and the target view with an inpainting region on the right. Qwen-Image-Edit and GPT-image-2 share the following insertion prompt. Seedance2-mini subsequently uses the generated endpoint image to produce a motion video. As in the question templates, braced placeholders denote task-specific values: \{object 1\} is the moved object, and \{moving group\} includes its rigid attachments.

\begin{AIbox}{Endpoint-Image Generation Prompt}
Use case: precise-object-edit. This is one two-panel image. The left panel is an appearance reference only. Use it to identify the same real \promptslot{object 1}, including its colors, materials, shape, proportions, texture, wear, and all rigidly attached components. Do not copy the left panel's room background, camera view, or unrelated objects. The right panel is the only destination scene and determines the final camera, perspective, lighting, focus, white balance, and stationary environment. Generate exactly one rigid moving object group (\promptslot{moving group}) only inside the white inpainting region in the right panel. The neutral low-detail noise in that region is an empty placeholder, not an appearance or material reference; replace it completely.
The inpainting region is a calibrated upper bound on the object's visible pixels. It fixes the object's destination position, maximum visible extent, scale, vertical height, and approximate silhouette. Fit the object inside that region at the destination camera's natural viewpoint. Do not move, recenter, enlarge, shrink, rotate arbitrarily, lower to the floor, raise, or duplicate it. Keep every listed attachment rigidly connected in the same relative arrangement as the left reference.
Make the inserted group a sharp photorealistic part of the right-panel photograph. Match perspective, illumination, shadows, reflections, focus, sensor noise, and occlusion boundaries. Preserve every foreground occluder and every unmasked pixel. Return the same two-panel layout. No cartoon, anime, illustration, CGI, plastic toy appearance, collage, isolated product shot, extra object, guide color, outline, arrow, text, or watermark.
\end{AIbox}

Qwen-Image-Edit additionally uses the following prompt to restore the background after removing the moving group from its original location.

\begin{AIbox}{Background-Restoration Prompt}
Photorealistic restoration of an empty background region in a real indoor RGB photograph. The white inpainting mask is the only editable area. Inside the white area, reconstruct only the stationary background that continues naturally from the immediately surrounding unmasked pixels, including the appropriate wall, floor, fixed worktop, baseboard, room corner, seam, or built-in support surface. Match the surrounding geometry, perspective, material, texture, lighting, shadows, focus, white balance, and sensor noise. The repaired area must look like an uninterrupted empty part of the original room. Keep every unmasked pixel, every stationary scene element, and the camera view unchanged. Do not generate any foreground object, furniture, appliance, container, or decoration in the mask. Produce a sharp seamless result with no blur, smear, repeated texture, patch boundary, text, or watermark.
\end{AIbox}

Seedance2-mini uses five reference images: @image1 is the initial view, @image2 identifies the moving group but is not a camera-route view, @image3 and @image4 show intermediate camera views, and @image5 is the generated endpoint image. The translation prompt is:

\begin{AIbox}{Video-Generation Prompt}
Create one continuous photorealistic four-second indoor tracking shot.
Use @image1 as the exact starting camera view and @image5 as the final camera view and final object state.
@image2 is a close identity reference for the complete \promptslot{moving group} only; it is not a camera-route waypoint and its cropped background must never appear as a shot.
Use @image3 and @image4, in that order, only as real or prepared intermediate camera-route references between @image1 and @image5.
Move exactly one \promptslot{moving group} by the authoritative world-space delta \promptslot{world displacement}, a distance of \promptslot{distance} meters. The question describes this as \promptslot{direction} in the frozen \promptslot{reference frame} frame.
Keep every member of the \promptslot{moving group} rigidly fixed relative to every other member and preserve identity, shape, scale, texture, lighting, and physical contact.
Remove the group from its old position as it moves; never leave a duplicate or ghost behind.
[\promptslot{occlusion constraint}]
Move the camera continuously from @image1 through @image3 and @image4 to @image5 with no unrelated detours.
Complete the object and camera motion during the first three seconds, then hold a stable view matching @image5 for the fourth second.
No cuts, crossfades, teleportation, mirroring, collage, text, arrows, colored marks, masks, bounding boxes, outlines, control lines, extra objects, or background collapse.
\end{AIbox}

For occlusion questions, the optional constraint reads: ``At the final view, preserve the requested pairwise occlusion relation: \{occlusion relation\}.'' For orbit questions, the translation sentence is replaced by: ``Move exactly one \{moving group\} in a \{angle\}-degree \{rotation direction\} orbit around the \{pivot object\} (pivot object ID \{pivot object ID\}). This is an orbit around the pivot, not an in-place spin: preserve the moving group's intrinsic world orientation throughout the orbit.''

\section{Case Study}
\label{app:case-study}

\begin{figure}[H]
    \centering
    \includegraphics[width=\textwidth]{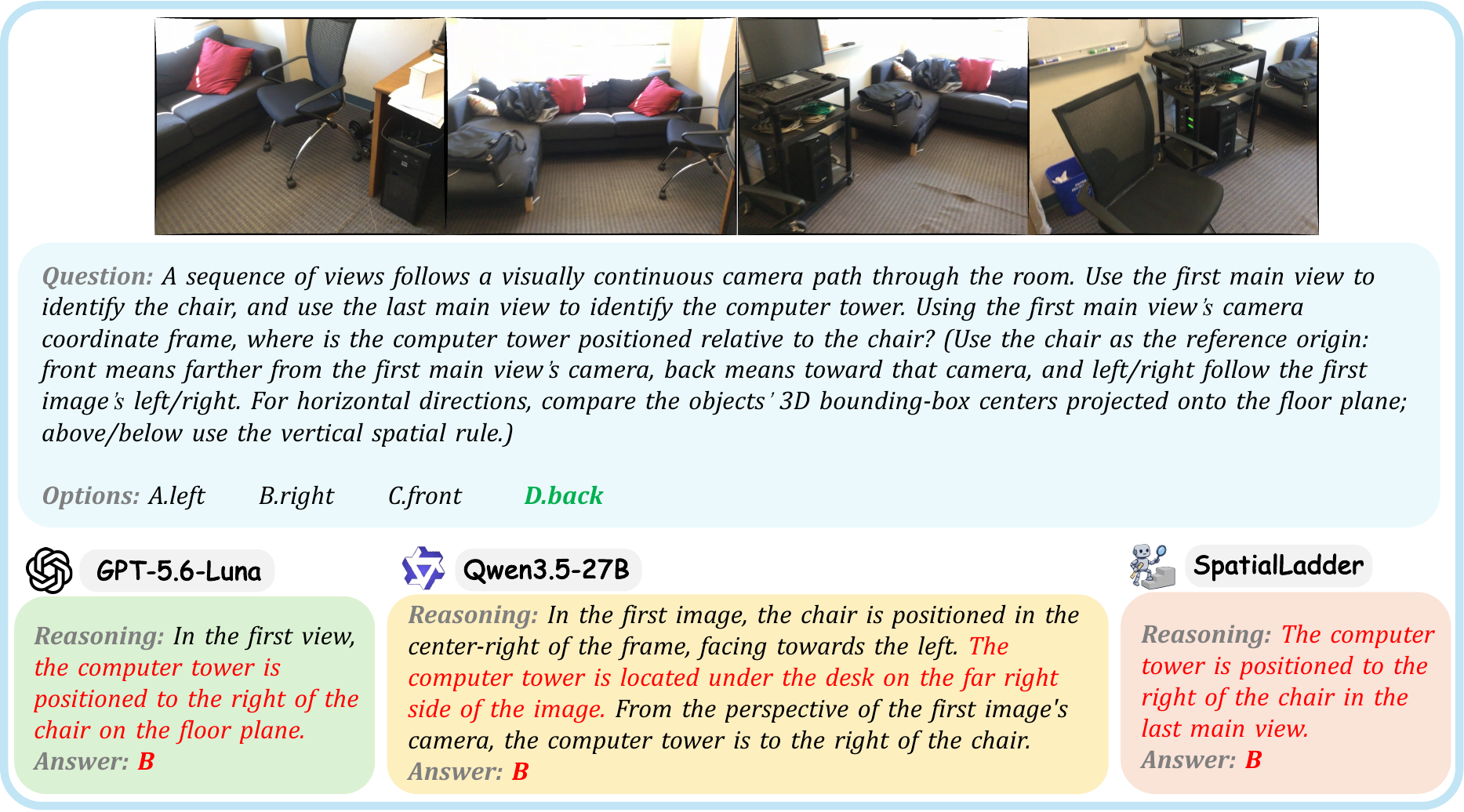}
    \caption{\textbf{L1 Camera Direction.} Rather than recovering the ground-plane layout and expressing it in the first main view's camera frame, models resort to image-space left--right positions or anchor on the last input view. Correct answers are shown in green; erroneous reasoning and predictions are highlighted in red.}
    \label{fig:case-camera-direction}
\end{figure}

\begin{figure}[H]
    \centering
    \includegraphics[width=\textwidth]{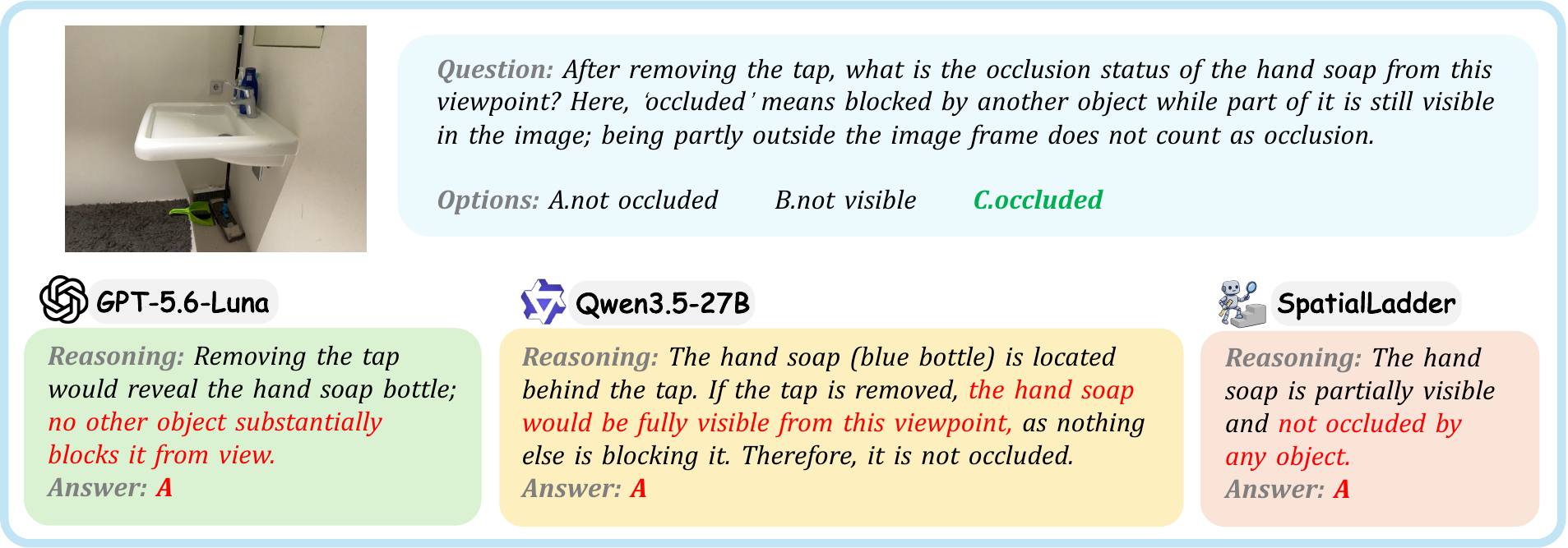}
    \caption{\textbf{L2 Remove Occlusion.} Models treat the removed tap as the sole occluder and conclude that the hand-soap bottle is now fully visible, failing to recompute visibility against the complete remaining scene. The case illustrates a broader pattern in which partial visibility is equated with the absence of occlusion. Correct answers are shown in green; erroneous reasoning and predictions are highlighted in red.}
    \label{fig:case-remove-occlusion}
\end{figure}

\begin{figure}[H]
    \centering
    \includegraphics[width=\textwidth]{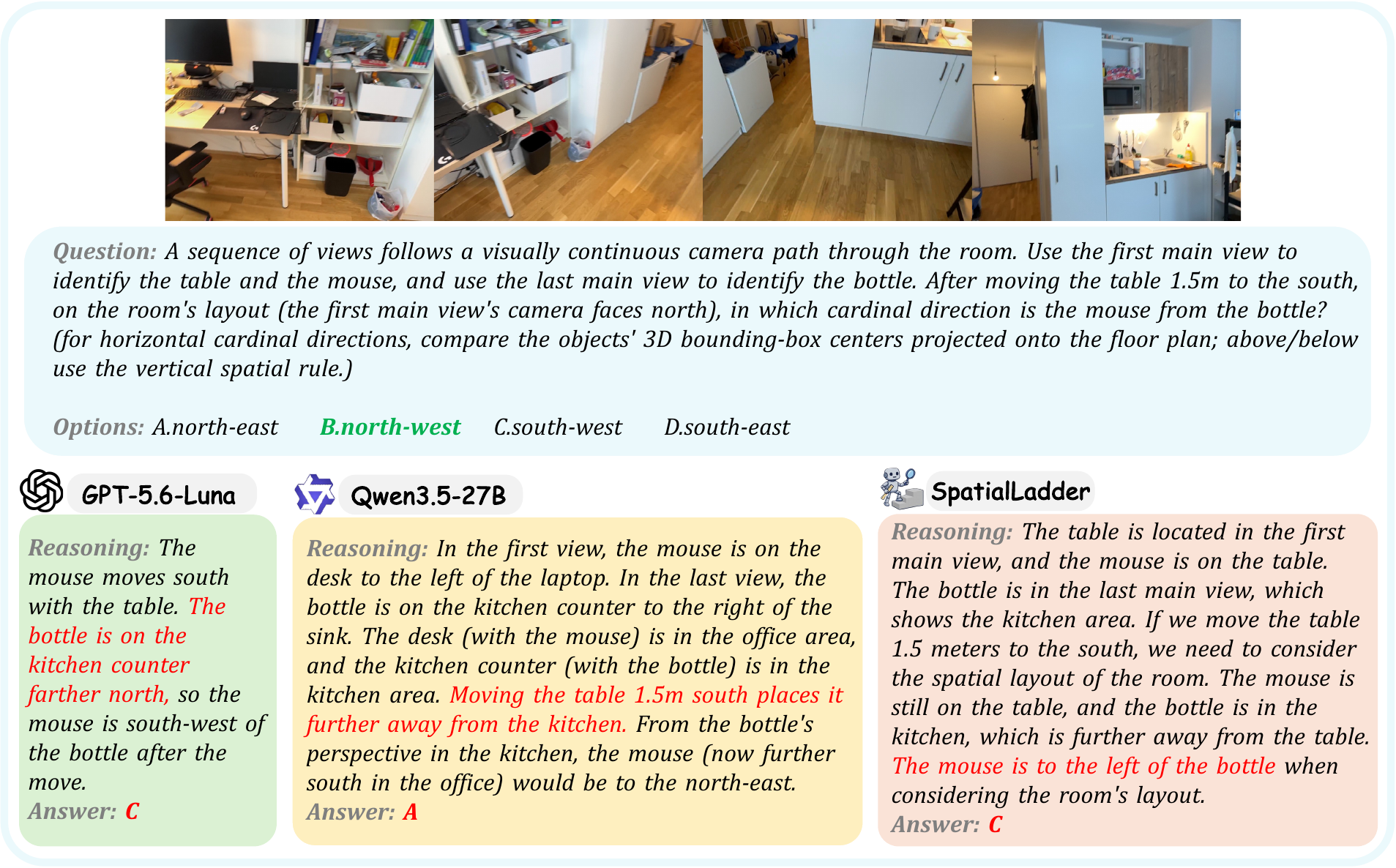}
    \caption{\textbf{L2 Move World Direction.} Models conflate the table's movement direction with the resulting direction of the mouse relative to the bottle, or substitute coarse room semantics for the required world-coordinate relation. The case further requires propagating the table's displacement to its rigidly attached mouse, an attachment dependency that models consistently overlook. Correct answers are shown in green; erroneous reasoning and predictions are highlighted in red.}
    \label{fig:case-move-world-direction}
\end{figure}

\begin{figure}[H]
    \centering
    \includegraphics[width=\textwidth]{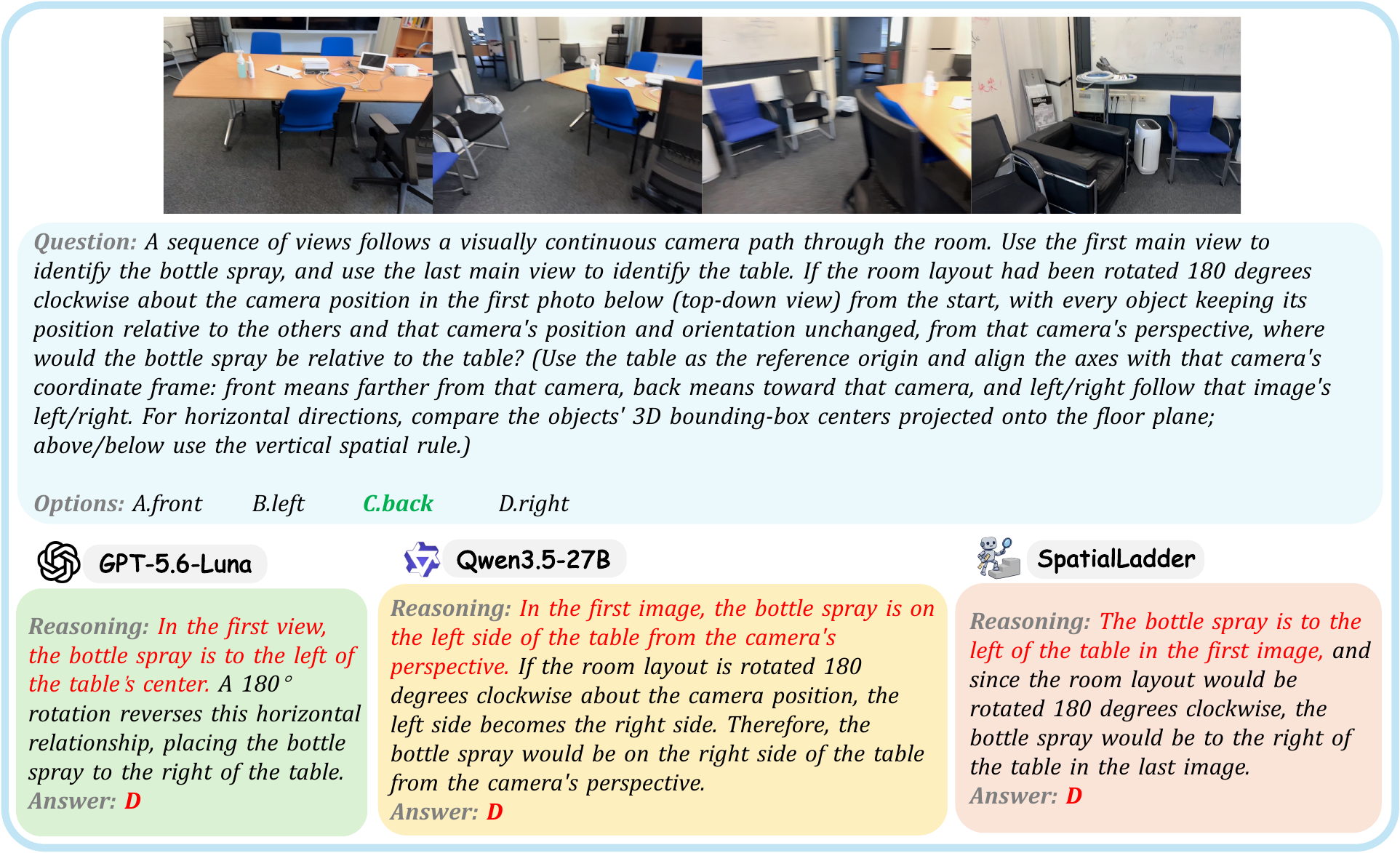}
    \caption{\textbf{L3 Rotate Camera Direction.} Models apply a superficial left--right reversal or reason from an unreformed input view rather than rotating the full scene layout and recomputing the queried relation in the fixed camera frame. The case illustrates how transformation reasoning degrades when it requires maintaining and manipulating a global scene representation. Correct answers are shown in green; erroneous reasoning and predictions are highlighted in red.}
    \label{fig:case-rotate-camera-direction}
\end{figure}

Figures~\ref{fig:case-camera-direction}--\ref{fig:case-rotate-camera-direction} present representative failure cases that reveal two intertwined limitations. The first is imperfect spatial perception: models construct only a shallow scene representation from the input views, anchoring on salient foreground objects or image-space cues rather than recovering consistent 3D structure. Depth ordering, ground-plane relations, and inter-object distances are frequently misestimated, and local observations across views are rarely integrated into a coherent whole. The second limitation is brittle transformation reasoning: even when the relevant objects are correctly located, models struggle to propagate an intervention through the scene. Movement direction is conflated with the resulting spatial relation, rigidly attached objects are left behind, and the post-intervention state is evaluated in the wrong reference frame or derived by a shallow 2D heuristic rather than genuine 3D recomputation. Removal questions expose a further failure mode in which visibility is only partially updated: models treat the deleted object as the sole occluder and equate a still-visible target with an unoccluded one. Taken together, these cases suggest that current vision-language models lack a persistent, manipulable scene representation, one that can be queried from an arbitrary viewpoint and updated under programmatic interventions, and that closing this gap is a prerequisite for reliable spatial reasoning in dynamic environments.

%% file: tables/table_app-task-composition.tex
\begin{table*}[h!]
    \centering
    \small
    \setlength{\tabcolsep}{4pt}
    \begin{adjustbox}{width=0.9\textwidth}
    \begin{tabular}{ccccccc}
        \toprule
        \textbf{Capability Level} & \textbf{Task Type} & \textbf{Relation} & \textbf{Reference Frame} & \textbf{ScanNet} & \textbf{ScanNet++} & \textbf{Total} \\
        \midrule
        \multirow{6}{*}{\textbf{L1: Static}} & Camera Direction & Direction & Camera-Centric & 72 & 192 & 264 \\
        & Object Direction & Direction & Object-Centric & 70 & 181 & 251 \\
        & World Direction & Direction & World-Centric & 66 & 197 & 263 \\
        & Distance & Distance & -- & 71 & 201 & 272 \\
        & Occlusion & Occlusion & -- & 173 & 48 & 221 \\
        & Attachment & Attachment & -- & 0 & 97 & 97 \\
        \midrule
        \multirow{7}{*}{\textbf{L2: Local}} & Move Camera Direction & Direction & Camera-Centric & 4 & 289 & 293 \\
        & Move Object Direction & Direction & Object-Centric & 4 & 252 & 256 \\
        & Rotate Object Direction & Direction & Object-Centric & 9 & 247 & 256 \\
        & Move World Direction & Direction & World-Centric & 2 & 262 & 264 \\
        & Move Distance & Distance & -- & 2 & 283 & 285 \\
        & Move Occlusion & Occlusion & -- & 2 & 137 & 139 \\
        & Remove Occlusion & Occlusion & -- & 37 & 179 & 216 \\
        \midrule
        \multirow{3}{*}{\textbf{L3: Global}} & Rotate Camera Direction & Direction & Camera-Centric & 71 & 192 & 263 \\
        & Rotate Object Direction & Direction & Object-Centric & 69 & 187 & 256 \\
        & Rotate World Direction & Direction & World-Centric & 66 & 200 & 266 \\
        \midrule
        \textbf{Total} & & & & 718 & 3,144 & 3,862 \\
        \bottomrule
    \end{tabular}
    \end{adjustbox}
    \caption{Composition of the 3,862-question evaluation set across 16 task types. The Relation column identifies the queried relation; Reference Frame applies only to direction tasks. Counts are broken down by ScanNet and ScanNet++.}
    \label{tab:app-task-composition}
\end{table*}

%% file: tables/table_app-question-templates-l1.tex
\begin{table}[htbp]
\centering
\small
\setlength{\tabcolsep}{5pt}
\renewcommand{\arraystretch}{1.0}
\newcommand{\qslot}[1]{\mbox{\textcolor{red}{\{#1\}}}}
\newcommand{\qtask}[1]{\parbox[c]{0.24\textwidth}{\raggedleft #1}}
\newcommand{\qvariants}[1]{\parbox[c]{0.67\textwidth}{\vspace*{2pt}\raggedright\itshape\begin{itemize}[leftmargin=1.1em,nosep,topsep=0pt,partopsep=0pt]#1\end{itemize}\vspace{2pt}}}
\begin{tabular}{@{}r|l@{}}
\textbf{Task} & \textbf{Question Template} \\
\hline
\qtask{Camera Direction} & \qvariants{
\item From the first main view's camera perspective, \qslot{object 1} is in which direction relative to \qslot{object 2}?
\item Using the first main view's camera coordinate frame, where is \qslot{object 1} positioned relative to \qslot{object 2}?
\item From the first main view's camera perspective, what is the spatial relationship of \qslot{object 1} to \qslot{object 2}?
} \\
\hline
\qtask{Object Direction} & \qvariants{
\item Imagine you are \qslot{object 1} and facing toward \qslot{object 2}. From your perspective, in which direction is \qslot{object 3}?
\item If you were \qslot{object 1}, looking toward \qslot{object 2}, where would \qslot{object 3} be?
} \\
\hline
\qtask{World Direction} & \qvariants{
\item The first main view's camera is facing \qslot{camera direction}. On the room's floor plan, in which cardinal direction is \qslot{object 1} from \qslot{object 2}?
\item In the first main view, the camera faces \qslot{camera direction}. Viewed from above on the room's layout, \qslot{object 1} is in which cardinal direction relative to \qslot{object 2}?
} \\
\hline
\qtask{Distance} & \qvariants{
\item What is the approximate shortest distance between \qslot{object 1} and \qslot{object 2}, measured from their closest points?
\item Measured from the closest points of each object, what is the approximate shortest distance between \qslot{object 1} and \qslot{object 2}?
\item Estimate the approximate shortest distance between \qslot{object 1} and \qslot{object 2}, measured from their closest points.
} \\
\hline
\qtask{Occlusion} & \qvariants{
\item What is the occlusion status of \qslot{object 1} in the current view?
\item From the current viewpoint, which best describes \qslot{object 1}: not occluded, occluded, or not visible?
\item In the current image, is \qslot{object 1} unoccluded, occluded by another object, or not visible?
} \\
\hline
\qtask{Attachment} & \qvariants{
\item Suppose \qslot{object 1} were moved to a different location. Which of the following objects would also be displaced from their current positions?
\item If \qslot{object 1} were relocated elsewhere in the room, which of the following objects would also change position? One or more options may be correct. Select all that apply.
\item Imagine \qslot{object 1} is moved to a new spot. Which of the following objects would also be displaced as a result? One or more options may be correct. Select all that apply.
} \\
\end{tabular}
\caption{L1 (Static Perception) templates for six task types querying original-scene relations or attachment structure. Red placeholders mark variable content.}\label{tab:app-question-templates-l1}
\end{table}

%% file: tables/table_app-question-templates-l2.tex
\begin{table}[htbp]
\centering
\small
\setlength{\tabcolsep}{5pt}
\renewcommand{\arraystretch}{1.0}
\newcommand{\qslot}[1]{\mbox{\textcolor{red}{\{#1\}}}}
\newcommand{\qtask}[1]{\parbox[c]{0.24\textwidth}{\raggedleft #1}}
\newcommand{\qvariants}[1]{\parbox[c]{0.67\textwidth}{\vspace*{2pt}\raggedright\itshape\begin{itemize}[leftmargin=1.1em,nosep,topsep=0pt,partopsep=0pt]#1\end{itemize}\vspace{2pt}}}
\begin{tabular}{@{}r|l@{}}
\textbf{Task} & \textbf{Question Template} \\
\hline
\qtask{Move Camera Direction} & \qvariants{
\item From the first main view's camera perspective, imagine moving \qslot{object 1} \qslot{direction} by \qslot{distance}. After this change, what is the relative position of \qslot{object 2} to \qslot{object 3}?
\item From the first main view's camera perspective, if we move \qslot{object 1} \qslot{direction} by \qslot{distance}, where is \qslot{object 2} relative to \qslot{object 3}?
\item From the first main view's camera perspective, suppose \qslot{object 1} is shifted \qslot{direction} by \qslot{distance}. After this change, what is the spatial relation of \qslot{object 2} with respect to \qslot{object 3}?
} \\
\hline
\qtask{Move Object Direction} & \qvariants{
\item In the initial scene, imagine \qslot{object 1} faces the camera in the first main view, and you are \qslot{object 2}, also facing that same camera. Freeze both objects' initial horizontal forward/right axes before anything moves. If \qslot{object 1} is moved \qslot{direction} by \qslot{distance} in its own frozen facing frame, from your separately frozen facing frame, in which horizontal direction is \qslot{object 3} from you afterward?
\item Suppose \qslot{object 1} and \qslot{object 2} each initially face the camera in the first main view. Keep each object's initial horizontal facing axes fixed throughout the motion. After moving \qslot{object 1} \qslot{direction} by \qslot{distance} in \qslot{object 1}'s frozen frame, where is \qslot{object 3} relative to \qslot{object 2} in \qslot{object 2}'s separately frozen frame?
} \\
\hline
\qtask{Rotate Object Direction} & \qvariants{
\item Imagine you are \qslot{object 1} and facing toward \qslot{object 2}. If \qslot{object 3} were moved along a \qslot{angle}-degree \qslot{rotation direction} (viewed from above) orbit around the center of \qslot{object 2} in the horizontal plane, without changing its own facing direction, from your perspective, in which direction would \qslot{object 4} be?
} \\
\hline
\qtask{Move World Direction} & \qvariants{
\item If \qslot{object 1} is moved \qslot{distance} to the \qslot{direction}, the camera in the first main view faces \qslot{camera direction}. On the floor plan, in which cardinal direction would \qslot{object 2} be from \qslot{object 3}?
\item After moving \qslot{object 1} \qslot{distance} to the \qslot{direction}, on the room's layout (the first main view's camera faces \qslot{camera direction}), in which cardinal direction is \qslot{object 2} from \qslot{object 3}?
} \\
\hline
\qtask{Move Distance} & \qvariants{
\item From the first main view's camera perspective, imagine moving \qslot{object 1} \qslot{direction} by \qslot{distance}. After this change, what is the approximate shortest distance between \qslot{object 2} and \qslot{object 3}, measured from their closest points?
\item From the first main view's camera perspective, if \qslot{object 1} is moved \qslot{direction} by \qslot{distance}, what is the approximate shortest distance between \qslot{object 2} and \qslot{object 3}, measured from their closest points?
} \\
\hline
\qtask{Move Occlusion} & \qvariants{
\item After moving \qslot{object 1} \qslot{direction} by \qslot{distance}, which best describes the pairwise occlusion relationship between \qslot{object 2} and \qslot{object 3} from the last main view's viewpoint?
\item From the last main view's viewpoint, after \qslot{object 1} is moved \qslot{direction} by \qslot{distance}, is \qslot{object 2} occluded by \qslot{object 3}, is \qslot{object 3} occluded by \qslot{object 2}, or does neither object occlude the other?
} \\
\hline
\qtask{Remove Occlusion} & \qvariants{
\item If \qslot{object 1} were removed from the scene, what would be the occlusion status of \qslot{object 2} from the current viewpoint?
\item Suppose \qslot{object 1} is taken away. Which best describes \qslot{object 2}: not occluded, occluded, or not visible?
\item After removing \qslot{object 1}, what is the occlusion status of \qslot{object 2} from this viewpoint?
} \\
\end{tabular}
\caption{L2 (Local Prediction) templates for seven task types involving object translation, orbit, or removal. Red placeholders mark variable content.}\label{tab:app-question-templates-l2}
\end{table}

%% file: tables/table_app-question-templates-l3.tex
\begin{table}[htbp]
\centering
\small
\setlength{\tabcolsep}{5pt}
\renewcommand{\arraystretch}{1.0}
\newcommand{\qslot}[1]{\mbox{\textcolor{red}{\{#1\}}}}
\newcommand{\qtask}[1]{\parbox[c]{0.24\textwidth}{\raggedleft #1}}
\newcommand{\qvariants}[1]{\parbox[c]{0.67\textwidth}{\vspace*{2pt}\raggedright\itshape\begin{itemize}[leftmargin=1.1em,nosep,topsep=0pt,partopsep=0pt]#1\end{itemize}\vspace{2pt}}}
\begin{tabular}{@{}r|l@{}}
\textbf{Task} & \textbf{Question Template} \\
\hline
\qtask{Rotate Camera Direction} & \qvariants{
\item Suppose this room had originally been designed with its orientation rotated \qslot{angle} degrees clockwise about the camera position in the first photo below (viewed from above), with every object keeping its position relative to the others. Observed from that same camera position and viewing direction (unchanged), in which direction is \qslot{object 1} relative to \qslot{object 2}?
\item If the room layout had been rotated \qslot{angle} degrees clockwise about the camera position in the first photo below (top-down view) from the start, with every object keeping its position relative to the others and that camera's position and orientation unchanged, from that camera's perspective, where would \qslot{object 1} be relative to \qslot{object 2}?
\item Imagine the room was originally built rotated \qslot{angle} degrees clockwise about the camera position in the first photo below (as seen from above). With every object keeping its position relative to the others and that same camera at its original pose, from that camera's perspective, what is the direction of \qslot{object 1} from \qslot{object 2}?
} \\
\hline
\qtask{Rotate Object Direction} & \qvariants{
\item Suppose this room had originally been oriented \qslot{angle} degrees clockwise about the camera position in the first anchor image (viewed from above), with every object keeping its position relative to the others. If you were \qslot{object 1} at its rotated position and kept facing the same horizontal direction that originally pointed from \qslot{object 1} toward \qslot{object 2}, in which direction would \qslot{object 3} be?
} \\
\hline
\qtask{Rotate World Direction} & \qvariants{
\item Imagine all furniture is rotated \qslot{angle} degrees clockwise about the camera position in the first photo below (viewed from above), with every object keeping its position relative to the others. That same camera, facing \qslot{camera direction}, remains in place. On the floor plan, in which cardinal direction is \qslot{object 1} from \qslot{object 2}?
} \\
\end{tabular}
\caption{L3 (Global Prediction) templates for three direction tasks after a counterfactual rotation of the full layout around a fixed camera. Red placeholders mark variable content.}\label{tab:app-question-templates-l3}
\end{table}

%% file: tables/table_sft-grpo-base-detailed.tex
\begin{table*}[h!]
    \centering
    \scriptsize
    \setlength{\tabcolsep}{1.2pt}
    \definecolor{StaticShade}{HTML}{F5FDF3}
    \definecolor{LocalShade}{HTML}{FFFDE9}
    \definecolor{GlobalShade}{HTML}{FEF1E5}
    \definecolor{ModelGroupShade}{HTML}{F3FBFF}
    \resizebox{\textwidth}{!}{%
    \begin{tabular}{r|c|*{19}{c}}
        &
        & \rotatebox{75}{Cam.\ Dir.} & \rotatebox{75}{Obj.\ Dir.} & \rotatebox{75}{World\ Dir.} & \rotatebox{75}{Dist.} & \rotatebox{75}{Occ.} & \rotatebox{75}{Att.} & \rotatebox{75}{\textbf{Avg.}}
        & \rotatebox{75}{Move\ Cam.\ Dir.} & \rotatebox{75}{Move\ Obj.\ Dir.} & \rotatebox{75}{Rotate\ Obj.\ Dir.} & \rotatebox{75}{Move\ World\ Dir.} & \rotatebox{75}{Move\ Dist.} & \rotatebox{75}{Move\ Occ.} & \rotatebox{75}{Remove\ Occ.} & \rotatebox{75}{\textbf{Avg.}}
        & \rotatebox{75}{Rotate\ Cam.\ Dir.} & \rotatebox{75}{Rotate\ Obj.\ Dir.} & \rotatebox{75}{Rotate\ World\ Dir.} & \rotatebox{75}{\textbf{Avg.}} \\
        \textbf{Model} & \textbf{Overall}
        & \multicolumn{7}{c}{\cellcolor{StaticShade}\textbf{L1: Static}}
        & \multicolumn{8}{c}{\cellcolor{LocalShade}\textbf{L2: Local}}
        & \multicolumn{4}{c}{\cellcolor{GlobalShade}\textbf{L3: Global}} \\
        \hline
        \rowcolor{ModelGroupShade}
        Qwen3-VL-4B & 34.0
        & 49.2 & 6.0 & 35.4 & 46.0 & 60.6 & 42.3 & 39.9
        & 36.9 & 28.5 & 24.2 & 29.5 & 31.9 & 51.8 & 32.9 & 33.7
        & 31.9 & 6.3 & 30.1 & 22.8 \\
        \textit{+ Mixed SFT} & 64.8
        & 84.8 & 90.8 & 65.0 & 62.9 & 80.5 & 74.2 & 76.4
        & 62.8 & 41.8 & 43.4 & 37.9 & 44.6 & 48.9 & 80.1 & 51.3
        & 77.6 & 80.5 & 61.3 & 73.1 \\
        \textit{+ Staged SFT} & 65.7
        & 88.6 & 88.8 & 61.2 & 68.0 & 80.5 & 73.2 & 76.7
        & 70.6 & 40.2 & 43.0 & 42.4 & 44.6 & 54.0 & 83.3 & 54.0
        & 79.1 & 79.7 & 53.8 & 70.8 \\
        \textit{+ GRPO} & 37.7
        & 53.4 & 7.6 & 38.8 & 29.4 & 63.3 & 49.5 & 40.3
        & 14.7 & 12.5 & 21.1 & 33.7 & 33.3 & 46.8 & 62.0 & 32.0
        & 42.6 & 63.7 & 30.8 & 45.7 \\
    \end{tabular}
    }
    \caption{Per-task accuracy (\%) of Qwen3-VL-4B and its trained variants on \dataname.}
    \vspace{-8pt}
    \label{tab:sft-grpo-base-detailed}
\end{table*}

%% file: tables/table_app-ood-types.tex
\begin{table*}[t]
    \centering
    \small
    \setlength{\tabcolsep}{5pt}
    \renewcommand{\arraystretch}{1.2}
    \begin{adjustbox}{width=0.85\textwidth}
    \begin{tabular}{>{\centering\arraybackslash}m{0.13\textwidth}|>{\centering\arraybackslash}m{0.07\textwidth}|>{\raggedright\arraybackslash}m{0.65\textwidth}}
        \textbf{Benchmark} & \textbf{Count} & \textbf{Included OOD Question Types} \\
        \hline
        MMSI & 274 & object\_motion and multi\_step\_reasoning \\
        \hline
        SPARTQA & 250 & ViewChg, SpImag\_OC, SpImag\_OC\_MV, SpImag\_OO, and SpImag\_OO\_MV \\
        \hline
        MindCube & 292 & Official dynamics/what-if slice: all sequence questions and rotation question families q2 and q3 \\
        \hline
        CLEVRER & 600 & Video temporal and causal reasoning \\
        \hline
        VSI & 150 & relative direction, relative distance, and route planning \\
        \hline
        DSI & 1,769 & object\_static\_camera, object\_moving\_camera, camera\_static\_scene, camera\_dynamic\_scene, object\_camera\_distance, and object\_camera\_orientation \\
    \end{tabular}
    \end{adjustbox}
    \caption{Question-type coverage and total counts of the six out-of-domain benchmarks.}
    \label{tab:app-ood-types}
\end{table*}